\pdfoutput=1
\documentclass[11pt]{article}

\usepackage[]{acl}

\usepackage{times}
\usepackage{latexsym}

\usepackage[T1]{fontenc}
\usepackage[utf8]{inputenc}

\usepackage{microtype}

\usepackage[linesnumbered,ruled,vlined]{algorithm2e}
\SetKwInput{KwInput}{Input}   

\usepackage[utf8]{inputenc} % allow utf-8 input
\usepackage[T1]{fontenc}    % use 8-bit T1 fonts
\usepackage{hyperref}       % hyperlinks
\usepackage{url}            % simple URL typesetting
\usepackage{amsfonts}       % blackboard math symbols
\usepackage{nicefrac}       % compact symbols for 1/2, etc.
\usepackage{microtype}      % microtypography
\usepackage{xcolor}         % colors

\usepackage{booktabs}
\usepackage{amsmath}
\usepackage{multirow}
\usepackage{graphicx}
\usepackage{enumitem}
\usepackage{subcaption}
\usepackage{caption}
\usepackage{rotating}
\usepackage{xcolor}

\newcommand{\mycomment}[3]{}

\newcommand{\ignore}[1]{}

\usepackage[normalem]{ulem}
\usepackage{amssymb}
\usepackage{pifont}% http://ctan.org/pkg/pifont
\usepackage{siunitx} 
\usepackage{multirow, makecell}
\usepackage{tabularx}% added for table design
\usepackage[all]{nowidow}

\definecolor{green}{RGB}{112, 173,71}
\definecolor{blue}{RGB}{68, 114,196}
\definecolor{orange}{RGB}{237, 125,49}
\definecolor{red}{RGB}{202, 54,49}
\definecolor{yellow}{RGB}{222,194, 142}
\usepackage{xspace}

\newcommand{\editor}{\textsc{EDITOR}\xspace}
\newcommand{\ourrollin}{\textsc{Editing}\xspace}
\newcommand{\cl}{\textsc{CL}\xspace}
\newcommand{\framework}{\textsc{Editing Curriculum}\xspace}  
\newcommand{\frameshort}{\textsc{EditCL}\xspace}  
\newcommand{\rollin}{roll-in\xspace}  
\newcommand{\rollout}{roll-out\xspace}  
\newcommand{\dual}{dual-path\xspace}

\newcommand{\dualtarget}{From Reference\xspace}
\newcommand{\dualsource}{From Input\xspace}

\newcommand{\source}{input\xspace}
\newcommand{\target}{output\xspace}

\newcommand{\nar}{\textsc{NAR}\xspace}
\newcommand{\ar}{\textsc{AR}\xspace}
\newcommand{\mt}{\textsc{MT}\xspace}
\newcommand{\ts}{\textsc{TS}\xspace}

\newcommand{\ari}{\textsc{ARI}\xspace}
\newcommand{\pcc}{\textsc{PCC}\xspace}
\newcommand{\sari}{\textsc{SARI}\xspace}

\newcommand{\cdf}{\textsc{CDF}\xspace}
\newcommand{\felix}{\textsc{FELIX}\xspace}
\newcommand{\rouge}{\textsc{ROUGE}\xspace}
\newcommand{\fmeasure}{\textsc{F1}\xspace}
\newcommand{\precision}{\textsc{P}\xspace}

\title{An Imitation Learning Curriculum for Text Editing  \\ with Non-Autoregressive Models}

\author{Sweta Agrawal \\
  Department of Computer Science \\
      University of Maryland \\
      {\tt sweagraw@cs.umd.edu} \\\And
  Marine Carpuat \\
  Department of Computer Science \\
  University of Maryland \\
  {\tt marine@cs.umd.edu} \\}

\begin{document}
\maketitle
\begin{abstract}
We propose a framework for training non-autoregressive sequence-to-sequence models for editing tasks, where the original \source sequence is iteratively edited to produce the output. We show that the imitation learning algorithms designed to train such models for machine translation introduces mismatches between training and inference that lead to undertraining and poor generalization in editing scenarios. We address this issue with two complementary strategies: 1) a roll-in policy that exposes the model to intermediate training sequences that it is more likely to encounter during inference, 2) a curriculum that presents easy-to-learn edit operations first, gradually increasing the difficulty of training samples as the model becomes competent. We show the efficacy of these strategies on two challenging English editing tasks: controllable text simplification and abstractive summarization. Our approach significantly improves output quality on both tasks and controls output complexity better on  the simplification task.
\end{abstract}

% \sa{Can we propose this work as a new non-autoregressive model for sequence editing, SeqEditor or TextEditor? While its true that we use Editor's operations and Levenshtein Distance as oracle, the training algorithm and the roll-in policy are novel and specific to the editing task. }

\section{Introduction}

% Motivation for edit based models for editing tasks
Neural sequence-to-sequence (seq2seq) models primarily developed and tested for machine translation (MT)  \citet{BahdanauChoBengio2015, vaswani2017attention, gu2018nonautoregressive} are increasingly used for other sequence transduction tasks. This paper focuses on \textit{editing tasks}, such as post-editing of MT output \cite{simard2007statistical}, style transfer \cite{jin2020deep}, or text simplification  \cite{chandrasekar1997automatic, XuCallison-BurchNapoles2015}, where systems directly edit the input sequence, instead of generating the output from scratch as in MT. As illustrated in Table~\ref{tab:example}, in these tasks, there might be substantial overlap in content between inputs and outputs, and also diverse rewrites, ranging from local substitutions to more complex restructuring. 

% Motivation for EDITOR, IL and CL
While dedicated architectures have been designed for these editing tasks, based on e.g., a multistep, tag-then-edit approach \cite{alva-manchego-etal-2017-learning, malmi2019encode, dong-etal-2019-editnts, mallinson-etal-2020-felix}, they can also be addressed with non-autoregressive (NAR) seq2seq models which generate their output by iteratively editing intermediate sequences \cite{lee2018deterministic, Gu2019Levenshtein,awasthi2019parallel, stern2019insertion, chan2020imputer}.
% \mc{Consider commeting out the editor cite for submission for anonymity, and add back later} \sa{Updated}
NAR models hold the promise of providing a more generic solution, where the model does not need to be tailored to a given editing task.

\begin{table}[t]
    \centering
        \renewcommand\tabularxcolumn[1]{m{#1}}% <-- added
        \renewcommand\arraystretch{1.3}
    \begin{tabularx}{\columnwidth}{*{1}{>{\arraybackslash}X}}
  \textbf{Original:}  The Mauritshuis museum is \textit{\color{orange} staging an exhibition focusing} on the 17th century self-portraits\textit{\color{blue}, highlighting} the similarities and the differences between \textit{\color{green} modern-day snapshots and historic works of art.} \\
  \textbf{Simplified:} The Mauritshuis museum is \textit{\color{orange} now set to open an exhibit} on the 17th century self-portraits\textit{\color{blue}. It shows} the similarities and differences between \textit{\color{green}modern photos and artworks.} \\
    \end{tabularx}  
    \caption{Text simplification is an editing task, where the output sequence overlaps with the input, while incorporating multiple rewrite types to restructure and simplify content.} \label{tab:example}
    % \vspace{-0.5cm}
\end{table}

% \sa{I tried editing the first sentence to make it more explicit but I am unsure if this rephrsing is better.}\mc{No, this is not addressing the comment. I did it myself in the interest of time.}
This work is centered on the hypothesis that training NAR models for editing tasks using the same strategy as for MT leads to a mismatch between train and test settings that limits their generalization ability and output quality. Specifically, the learning algorithms designed for MT are aligned with inference strategies that generate \target from an empty initial sequence. By contrast, in sequence editing tasks, the inference step is initialized instead with the original \source sequence. In addition, since editing samples might range from limited lexical substitutions to more thorough rewrites,  training samples cover a wide range of edit distances. During training, the loss can thus be dominated by the more distant samples leading to undertrained models and poor generalization. By contrast, the distance between input and output samples in MT is more uniform, since it always involves at least lexical translation of the input tokens.

To address these issues, we introduce a new training framework, \textit{\framework}, which dynamically exposes the model to more relevant edit actions during training and exploits the full spectrum of available training samples more effectively. First, we design a new roll-in strategy, \textit{\ourrollin \rollin}, that exposes the model to intermediate sequences that it is more likely to encounter during inference. Second, we introduce a training \textsc{curriculum} to expose the model to training samples in order of increasing edit distance, thus gradually increasing the complexity of oracle edit operations that the model learns to imitate.

We show that our approach improves the quality of outputs on two challenging English text editing tasks: controllable text simplification (\ts) and abstractive summarization. It also improves the degree of \ts control by generating simplified outputs that match the target reading grade level better than the baselines. We conduct an extensive analysis which supports our hypothesis, and show that the sequences generated by our training policy improve exploration during training and are easier to learn from, leading to better generalization across samples with varying edit distances. Training with curriculum further improves output quality.

% \input{tables/meta_compare_policies}

% Section 2-3: Method

% For placing

\begin{table*}[t]
    \centering
     \setlength\tabcolsep{3pt}
    \scalebox{0.78}{
    \begin{tabular}{lllr}
   & \multicolumn{1}{c}{\textsc{Operations}}  & \multicolumn{1}{c}{Roll-In} &  \multicolumn{1}{c}{\textsc{Roll-in Policies}}   \\
   \midrule
    \addlinespace[0.2cm]
     \citet{Gu2019Levenshtein} & Insertion, Deletion & Mixed &   $y'=\{ \mathcal{E}(y^*, \tilde{d}), \tilde{d} \sim \pi_{rnd}$ \}\\
    & & &   $y_{ins} = \{ y'$ if  $u<\alpha$ else $\mathcal{E}(y^s, d^*), d^* \sim \pi^*_{del} \}  $ \\
   & & &   $y_{del} = \{ y^s$ if $u<\beta$ else $\mathcal{E}(\mathcal{E}(y_{ins}, p^*), \tilde{t}), p^* \sim \pi^*_{plh}, \tilde{t} \sim \pi_{ins} \} $\\ 
         \addlinespace[0.1cm]
   
   \citet{stern2019insertion} & Insertion & Expert  &   $y_{ins} = \{ \mathcal{E}(y^*, \tilde{d}), \tilde{d} \sim \pi_{rnd}$ \} \\
   
    \addlinespace[0.1cm]
   
    \citet{ghazvininejad2019mask} & Substitution & Expert &   $y_{sub} = \{ \mathcal{E}(y^*, \tilde{m}), \tilde{m} \sim \pi_{mask}$ \} \\
%   \citet{ghazvininejad2019mask} & Insertion & Expert &   $y_{ins} = \{ \mathcal{E}(y^*, \tilde{d}), \tilde{d} \sim \pi_{rnd}$ \} \\
  
     \addlinespace[0.1cm]
     
    %  \citet{saharia-etal-2020-non} & Insertion & \small {(Offline)} Learned & $y_{ins} = \{ \mathcal{E}(y, \tilde{d}), \tilde{d} \sim \pi_{rnd}$ \}  \\
     \citet{saharia-etal-2020-non} & Substitution & \small{(Offline)}  Learned & $y_{sub} = \{ \mathcal{E}(y, \tilde{m}), \tilde{m} \sim \pi_{mask}$ \}  \\

     \addlinespace[0.1cm]
     
    %  \citet{qian2020glancing} & Insertion & Expert & $y_{ins} = \{ \mathcal{E}(y, \tilde{d}), \tilde{d} \sim \pi_{rnd}$ \}
     
      \citet{qian2020glancing} & Substitution & Expert & $y_{sub} = \{ \mathcal{E}(y, \tilde{m}), \tilde{m} \sim \pi_{mask}$ \} \\
      
          \addlinespace[0.1cm]
      
    \citet{xu2021editor} & Insertion, Reposition & Learned &  $y' =\{ \mathcal{E} (\mathcal{E}(y^*, \tilde{d}), \tilde{p}) , \tilde{d} \sim \pi_{rnd}, \tilde{p} \sim \pi_{per} \} $ \\
       &  \small (including deletions) & &   $y_{ins} = \{ y'$ if  $u<\alpha$ else $\mathcal{E}(y, r), r \sim \pi_{rps} \}  $ \\
      & & &    $y_{rps} = \{ y'$ if  $u<\beta$ else $\mathcal{E}(\mathcal{E}(y, p^*), \tilde{t}), p^* \sim 
      \pi^*_{plh}, \tilde{t} \sim \pi_{ins} \}  $ \\
    
    \end{tabular}}
    \caption{Training Policies and Edit Operations performed by different NAR models: $y^s$: original \source sequence, $y^*$: \target sequence,  $y$: model generated variant of reference sequence, $\pi_{rnd}$/ $\pi_{masks}$ drops/masks random words from $y^*$ according to a distribution (e.g. uniform, bernoulli, etc.),  $\pi_{p}$ generates a permutation,  $u: \sim Uniform[0, 1]$, $\pi_{ins}, \pi_{plh}, \pi_{del}, \pi_{rps}$ are insertion, placeholder prediction, deletion and reposition policies. }
    \vspace{-0.5cm}
    % \sa{Double checked with Weijia if the interpretation is correct, she seems to agree. Is the table reasonable?}\mc{Need a rationale for ordering the models. It could be as simple as chronological order, but the papers need to be organized in some way.} \sa{Yes, updated. I initially intended it to be that way but messed up one paper.}
    \label{tab:policy_compare}
\end{table*}
\section{Background} \label{sec:background}

 \paragraph{Model} \nar edit-based models \cite{chan2020imputer, Gu2019Levenshtein, stern2019insertion, xu2021editor} cast sequence editing as an iterative sequence refinement problem modeled by a Markov Decision Process $\left(\mathcal{Y}, \mathcal{A}, \mathcal{E}, \mathcal{R}, \boldsymbol{y}^{0}\right)$. A state $y=(y_1, y_2, ..., y_{L}) \in \mathcal{Y}$ is a sequence of tokens where each $y_i$ represents a token from the vocabulary $\mathcal{V}$, $L$ is the sequence length and $y^0 \in \mathcal{Y}$ is the initial sequence to be refined, using actions drawn from the set $\mathcal{A}$. The reward $\mathcal{R}$ is based on the distance $\mathcal{D}$ between the generated \target and the reference sequence $y^{*} \in \mathcal{Y}$: $\mathcal{R}(y) = -\mathcal{D}(y, y^*)$. At each decoding iteration, the model takes an input $y$, chooses an action $a \in \mathcal{A}$ to refine the sequence using a policy $\pi$, resulting in state $\mathcal{E} (y, a)$. 

Models differ based on the nature of edit actions used and support different operations such as insertion, deletion, reposition and substitution. We select the operations from the \editor model based on its competitive performance on constrained decoding tasks that require editing non-empty initial sequences \citep{xu2021editor}. It is a Transformer model that uses two types of actions or edits on sequences, $y$: 

\begin{enumerate}
    \item The \textbf{reposition} operation, modeled by $\pi_{rps}$, predicts the new position of each token in the input sequence. For each input position, the reposition policy predicts a value $r$ that corresponds to the index of the input token to be placed at the position and $0$ if the input token is to be deleted. 
    \item The \textbf{insertion} operation has two components: \textit{placeholder prediction}, $\pi_{plh}$ that predicts the number of placeholders to be inserted and \textit{token prediction}, $\pi_{ins}$ that generates the actual \target tokens for each placeholder.
\end{enumerate}
 At each decoding iteration, the model applies an action $a$ that consists of a reposition and an insertion operation. This refinement process is repeated until two consecutive decoding iterations return the same output \cite{Gu2019Levenshtein}, or a preset maximum number of them is reached \cite{lee2018deterministic, ghazvininejad2019mask}. 

\begin{figure}[htb!]
\centering
  \includegraphics[width=0.7\linewidth]{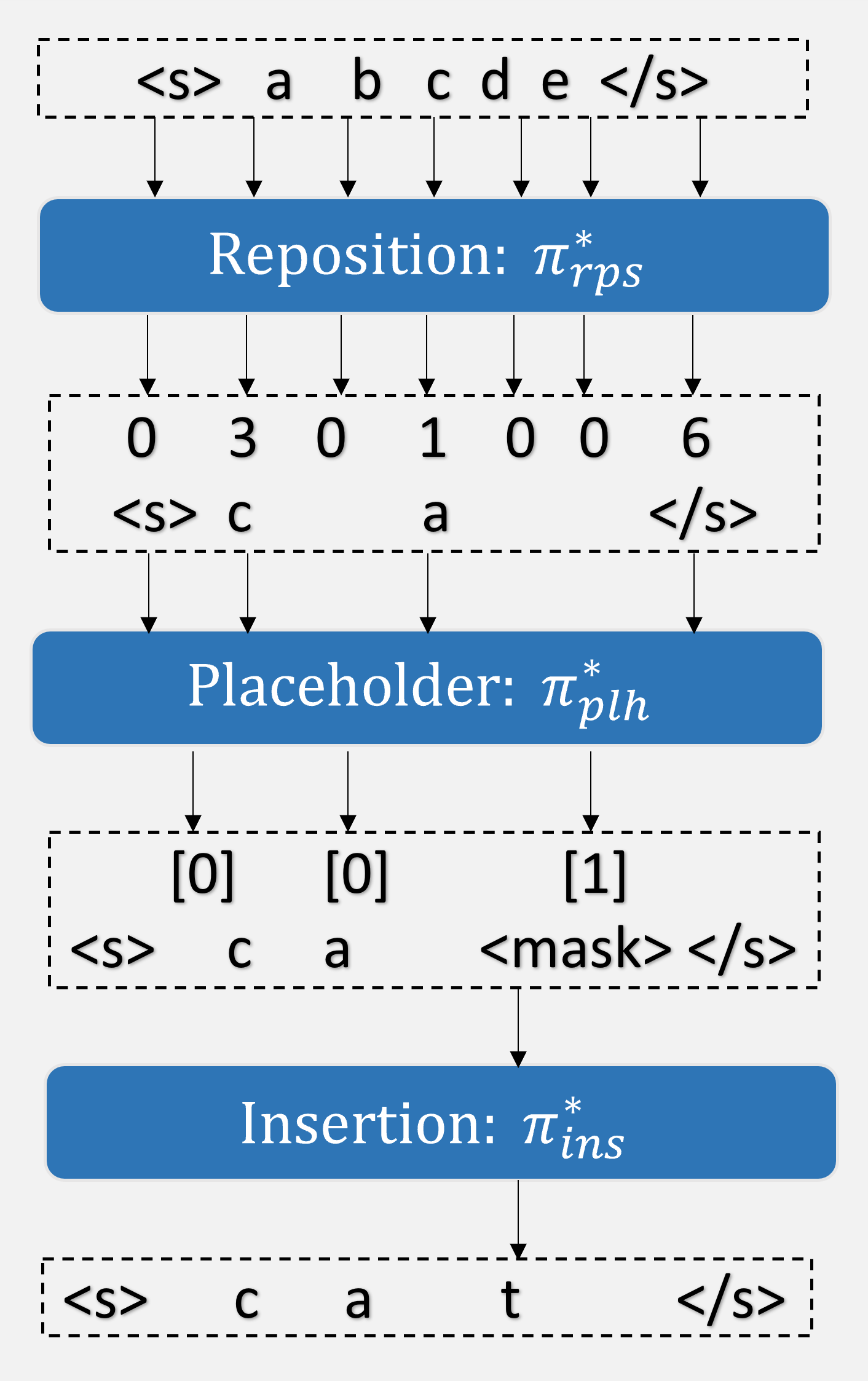}
    \caption{ One refinement iteration for the \source sequence: "a b c d e" using the operations generated by the Levenshtein Edit Distance Algorithm. }  \label{fig:example_editor}
     \vspace{-0.5cm}
\end{figure}

\paragraph{Training} \nar models are typically trained via imitation learning that uses a \rollin policy and a roll-out policy. The \rollin policy is used to generate the sequences that the model learns to refine from. A roll-out policy is then used to estimate the cost-to-go from the generated \rollin sequences to the desired \target sequences. The cost-to-go is calculated by comparing the model actions to oracle demonstrations. We summarize the policies of various \nar models proposed for \mt in Table~\ref{tab:policy_compare}.

For \editor, the roll-in sequences for the reposition (or insertion) module are stochastic mixtures (parameterized by $\alpha$ or $\beta$) of the output of the insertion (or reposition) module or a noised version of the \target sequence. The oracle is the Levenshtein edit distance \cite{Gu2019Levenshtein}.  The noisy sequence is generated by applying random word dropping \cite{Gu2019Levenshtein} and random word shuffle \cite{lample2018unsupervised} with a probability of 0.5 and maximum shuffle distance of 3. Figure~\ref{fig:example_editor} shows an example instantiation of the edit actions generated by the Levenshtein Edit Distance to transform the original \source sequence (``a b c d e'') to the \target sequence (``c a t'').  In this example, the oracle action is to delete the tokens [``b'', ``d'', ``e''], reposition ``a'' and ``c'' and insert ``t'' at the appropriate position.  The reposition and the insertion modules are trained in a supervised fashion to predict these oracle operations during training.

\section{Our Approach: \framework}  \label{sec:approach}

%While most prior efforts have been towards designing  \nar models for \mt, we adapt them for sequence editing tasks and hypothesize that training for these tasks with the same policy as in \mt could be sub-optimal.
To tailor training to editing tasks, we propose to modify the roll-in policy to better match the intermediate sequences encountered at inference, and introduce a curriculum to increase the difficulty of oracle actions learned throughout training. 

% \mc{Noisy Expert Roll-in => Roll-in for Editing? any other ideas?} \sa{Edit Roll-in?}
 \paragraph{\ourrollin Roll-in} Sequences generated using the \rollin policy control the search space explored during training. Those sequences should therefore be representative of the intermediate sequences generated at inference time \cite{pmlr-v9-ross10a}.  While typically, the \rollin policy is a stochastic mixture of the model and the expert demonstrations as described above, the noise incurred early on due to the large difference between the expert demonstration and the learner’s policy actions may hurt overall performance \cite{brantley-etal-2019-non, he2012imitation, leblond2018searnn}. As we will see (\S\ref{sec:main_results}), this is what happens on editing tasks when training the model to imitate experts using learned \rollin sequences. At the same time, rolling in with expert demonstrations raises its own issues, as it can limit the exploration of the search space.
 % For placing

% Algorithm
\begin{algorithm*}

\KwInput{Dataset, $D= \{y, y^*\}_{i=1}^M$, difficulty scoring function, $d$, and competence function, $c$.}
Compute the difficulty, $d(s_i)$, for each $s_i = \{y_i, y^*_{i}\} \in D $.\\

Compute the cumulative density function (CDF) of the difficulty scores. This results in one difficulty CDF score per sample, $\tilde{d}(s_i) \in [0, 1]$
% \mc{How is $s_i$ defined exactly?} \sa{Added}.

Initialize $\pi_{rps}$ and $\pi_{ins}$. \\

   \For{training step $t=1...T$}
   {
   		 Compute the competence value, $c(t)$. \\
   		 Create training dataset from by selecting all samples, $B_{t}$ using $s_i \in D $, such that $\tilde{d}(s_i) \leq c(t)$. \\
   		 \For{i in $1 .. |B_{t}|$}
           {
            Generate roll-in sequences:  \\  
            ~~$y_{rps} = noise(y^s) $ \\
            ~~$y_{ins} = \mathcal{E}(y_{rps}, r^*), r^* \sim \pi^*_{rps}$ \\
            Train  $\pi_{rps}$ and $\pi_{ins}$ on $y_{rps}$ and $y_{ins}$ minimizing cost-to-go to $y^*$. \\
           }
   }
   Return \textbf{best} $\pi_{rps}$ and $\pi_{ins}$ evaluated on validation set.
	\caption{Our proposed framework: \framework} \label{algorithm}
\end{algorithm*}

% \vspace{-1cm}

\begin{figure}[htb!]
\centering
  \includegraphics[width=\linewidth]{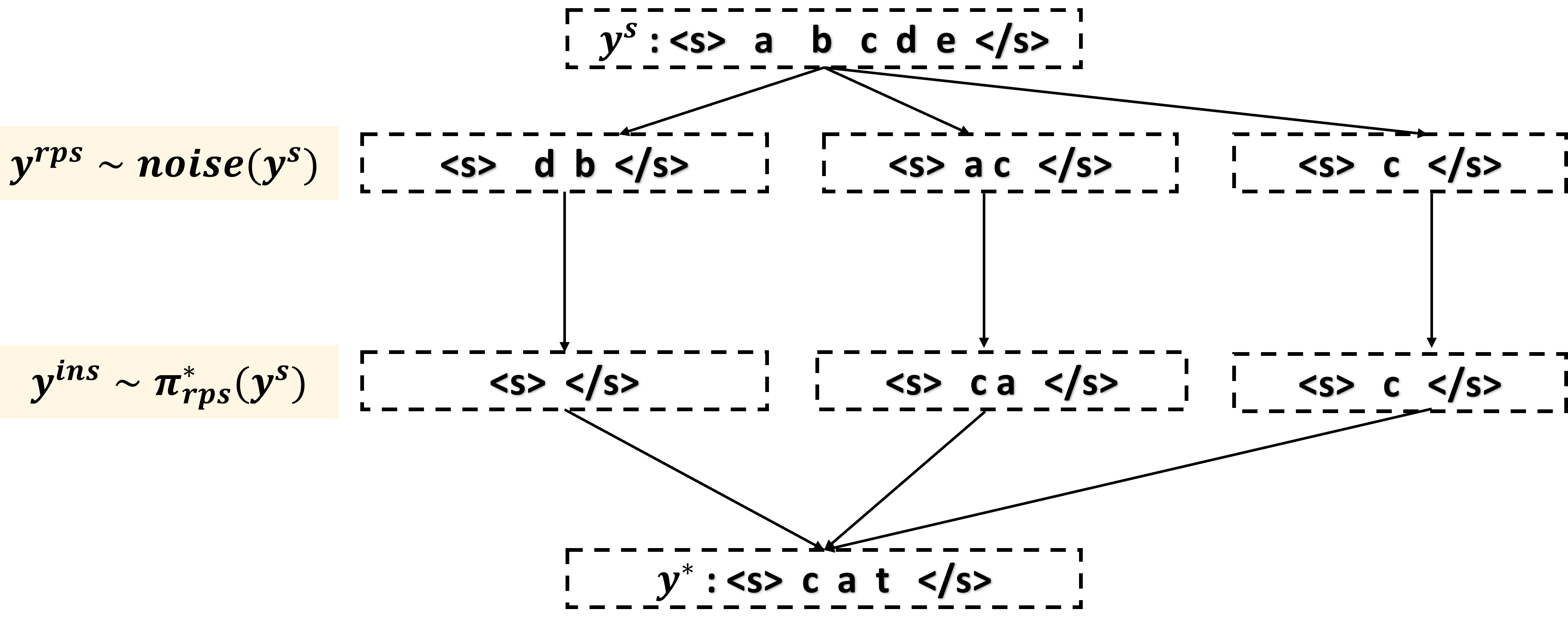}
    \caption{Example \rollin sequences for the reposition and the insertion modules: The same initial \source sequence ($y^s$) can enable the model to learn to generate the reference output ($y^*$)  using different edit operations from its noised version. }  \label{fig:example_rollin}
\end{figure}
 
Motivated by these observations, we propose a new policy, \textit{\ourrollin}, that allows exploration by injecting noise to the \source sequence to generate new intermediate sequences for training. This lets the model learn to fix errors without deviating from learning the task at hand. Figure~\ref{fig:example_rollin} shows an example of intermediate sequences generated by our proposed \rollin policy. Different intermediate sequences encourage the model to learn different reposition and insertion edit operations starting from the same \source sequence, hence enabling exploration. We modify the \rollin policies to be aligned with the editing inference process, where the reposition operation is followed by insertion on the original \source sequence: 

% \sa{Updated below to incorporate reviewers suggestions}
\begin{itemize}[]
    \item The \rollin sequence for training the reposition module, $\pi_{rps}$, is generated by applying noise to the original source sequence $y^s$, i.e. $y_{rps} = noise(y^s) = \{ \mathcal{E} (\mathcal{E}(y^s, \tilde{d}), \tilde{p}) , \tilde{d} \sim \pi_{rnd}, \tilde{p} \sim \pi_{per} \} $.  Unlike \editor, the random word dropping ($\tilde{d} \sim \pi_{rnd}$) and the word shuffling ($ \tilde{p} \sim \pi_{per}$) are applied to the original \source sequence instead of the \target sequence. This aligns the training with the inference scenario where the model edits an original \source sequence instead of generating an output from scratch.
    
    \item The \rollin sequence for training the insertion module, $\pi_{ins}$ is an intermediate sequence generated by applying the expert reposition policy to  $y_{rps}$, i.e. $y_{ins} = \{\mathcal{E}(y_{rps}, r^*), r^* \sim \pi^*_{rps} \} $. The expert reposition policy corresponds to the deletion and reposition actions derived by using the levenshtein edit distance algorithm between the noisy \source sequence, $noise(y^s)$ and the target sequence, $y^*$.
\end{itemize}

\paragraph{Curriculum controlled \rollout}
To prevent undertraining when samples with large edit distances overwhelm the loss, we use a curriculum to expose the model to easy-to-learn actions first, then gradually increase the difficulty of the edit-operations performed as the learner becomes more competent. Prior work on curriculum learning (\cl) does not agree on standard measures of sample difficulty for seq2seq tasks \cite{ kumar-etal-2019-reinforcement, yao2021learning, zhang2018empirical, zhou-etal-2020-uncertainty} or apply \cl for the different problem of shifting the training of a Transformer model from \textsc{AR} to \nar regimes \cite{guo2020fine, liutask}. By contrast, in our settings, the Levenshtein distance provides a measure of difficulty that directly aligns with the model design and the training oracle.
% \sa{need to add details on how its used or move to next paragraph} 
% Given this ranking, the learning schedule is governed by the competence function $c(t) \in(0,1]$ of \cite{platanios2019competence}. We use their proposed root-based pacing with default settings

% \sa{Updated to include more details and restructured}
\paragraph{Resulting Algorithm} 
Given a training dataset $\mathcal{D}=\{y^s, y^*\}_{i=1}^M$ consisting of $\mathcal{M}$ samples, the difficulty score $d(s_i)$ for each sample $s_i=\{y_i^s, y_i^*\} \in \mathcal{D}$ is measured by the Levenshtein Distance between the \source and the \target sequence. The cumulative density function (\cdf) of the difficulty scores results in one difficulty \cdf score per sample, $\tilde{d}(s_i)$. At each training step $t$,  we estimate the progress made by the learner by computing the competence of the model $c(t) \in(0,1]$ as follows:  
$$c_{\text {sqrt }}(t) =\min \left(1, \sqrt{t \frac{1-c_{0}^{2}}{\lambda_{t}}+c_{0}^{2}}\right)$$
where, $\lambda_t$ defines the length of the curriculum\footnote{ We set the curriculum length to $5$K for our experiments.}; $c_0 = 0.1$ as in \newcite{platanios2019competence}. 

Based on this competence value $c(t)$, the model is then trained on all the samples whose difficulty as measured by the Levenshtein distance between the \source and the \target sequence is lower than that competence value, i.e. $\tilde{d}(s_i) \le c(t)$. The resulting algorithm is also shown in Algorithm~\ref{algorithm}.

%

% Section 4: Experiment-setup 

\section{Experimental Settings}

We evaluate our approach on Controllable Simplification and Abstractive Summarization, two challenging sequence editing tasks that are motivated by real world information access needs. They are challenging because they require learning to perform a wide range of rewrites (from local substitution to sentence restructuring).

\subsection{Controllable Simplification}
\paragraph{Task Definition} Given a complex text and a target grade level, the goal is to generate a simplified output that is appropriate for the desired grade level. The type of operations performed across different grade levels span sentence splitting, paraphrasing, deletion, content elaboration and substitution. 

\paragraph{Data}  We use English Newsela samples as extracted by \citet{agrawal2019controlling} with $470$k/$2$k/$19$k for training, development and test sets respectively. Grade side-constraints are defined using a distinct special token for each grade level (from 2 to 12) and are introduced as side constraints for both the \source and the \target grade levels \citet{ScartonSpecia2018}. 

\paragraph{Evaluation Metrics} 
We automatically evaluate truecased detokenized system outputs using: \textbf{SARI} \cite{XuNapolesPavlickChenCallison-Burch2016}, which measures the lexical simplicity based on the n-grams kept, added, and deleted by the system relative to the \source and the \target sequence. It computes the \fmeasure score for the n-grams that are added (\textbf{add-\fmeasure}). The model’s deletion capability is measured by the \fmeasure score for n-grams that are kept (\textbf{keep-\fmeasure}) and precision for the n-grams that are deleted (\textbf{del-\precision}) \footnote{\url{https://github.com/cocoxu/simplification}}; \textbf{Pearson’s correlation coefficient (PCC)} between the complexity of the system and reference outputs as measured by Automatic Readability Index (ARI) \cite{SenterSmith1967} and \textbf{ARI-Accuracy} \cite{HeilmanCollinsEskenazi2008} representing the percentage of sentences where the system output grade level is within 1 grade of the reference text according to the ARI.

\subsection{Abstractive Summarization}

\paragraph{Task} Given a short paragraph (one or two sentences on average), the goal is to generate a concise summary that captures the salient ideas of the source text. It contains heavy deletions
with moderate amounts of substitutions and frequent shifts caused by re-orderings.

\paragraph{Data} We use the dataset from \citet{toutanova-etal-2016-dataset},
which contains $6$K short input texts, with upto 5 summaries each. We use the same split as provided by the authors with $4937$/$448$/$786$ unique input texts in the training, development and test sets respectively. The human experts were allowed to insert new words and reorder parts of the sentence when generating the summary, which makes this dataset particularly suited for abstractive summarization models.

\paragraph{Evaluation Metrics} We automatically evaluate truecased detokenized system outputs using:  \textbf{Rouge-L}\footnote{\url{https://github.com/pltrdy/rouge}}\cite{lin-2004-rouge}. Even though it is not a summarization metric, we also report \textbf{SARI} to track the nature and type of edit operations performed. Given multiple references for each input text, we define the corpus level score as the arithmetic mean of automated metrics at the instance level, which is further averaged across the multiple references.

\subsection{Model configurations}

\paragraph{Data Preprocessing} We pre-process all data using Moses tools for normalization, and truecasing. We apply subword segmentation with a joint \source-\target byte pair encoding model with $32,000$ operations. We use \ari to compute the \source grade level at the inference time.

\paragraph{Architecture} We adopt the base Transformer architecture \cite{vaswani2017attention} with $d_{model}$ = $512$, $d_{hidden}$ = $2048$, $n_{heads}$ = $8$, $n_{layers}$ = $6$, and $p_{dropout}$ = $0.1$ for all our models. We add dropout to embeddings ($0.1$) and label smoothing ($0.1$). The base \editor model is trained using Adam with initial learning rate of $0.0005$ and a batch size of $16,000$ tokens. The model is further finetuned on the editing task with a learning rate of $0.0001$. We train all our models on two GeForce GTX 1080Ti GPUs. The average training time for a single seed of AR model is $\sim$8-9 hrs and for the \editor model is $\sim$20-22 hrs. Fine-tuning \editor takes additional 5-6 hrs. Training stops after $8$ checkpoints without improvement of validation perplexity. All models are implemented using the Fairseq toolkit.

\paragraph{Models}  We compare our proposed approaches against the following models trained from scratch in controlled conditions:
1) {\ttfamily \textbf{AR}} is a auto-regressive (AR) transformer model \cite{ScartonSpecia2018}. 2) We train \editor with the \dual roll-in policy as in \citet{xu2021editor}, refered to as {\ttfamily \textbf{\dualtarget}}. We fine-tune EDITOR with the following policy variants: 3) { \ttfamily \textbf{\dualsource}} replaces the reference with the \source for generating the initial sequence as in \citet{agrawal2021non}. 4) {\ttfamily \textbf{Editing}} is our proposed roll-in policy. 5)  {\ttfamily \textbf{Editing Curriculum}}, \frameshort, refers to our approach as described in ~\S\ref{sec:approach}.  
During inference, we start from the \source sequence ($y^s$), which is refined iteratively by applying a sequence of actions, as described in ~\S\ref{sec:background} until 1) the output
sequences from two consecutive iterations are the
same, or 2) the maximum number of decoding steps ($N=10$) is reached. The edit distance between two sequences is measured by the Levenshtein edit distance \cite{levenshtein1966binary}.

%

% Section 5: results 

% for table positioning
% simplification main
\begin{table*}[t]
\centering
 \setlength\tabcolsep{4pt}
\scalebox{0.80}{
\begin{tabular}{lrrrrcrrcrr}
 \toprule
  \multirow{2}{*}{\textbf{Model}} & \multicolumn{4}{c}{\bf{SARI}}  & & \multicolumn{2}{c}{\bf{ARI-based}} & & \multirow{2}{*}{\shortstack{\textbf{Training} \\ \textbf{Updates}}} &  \multirow{2}{*}{ \shortstack{\textbf{Inference } \\\textbf{ action/sample }}}\\

 \cmidrule{2-5} \cmidrule{7-8} 
 &   \textbf{keep-F1} & \textbf{add-F1} & \textbf{del-P} & \textbf{combined} & & \textbf{PCC} &  \textbf{\% ARI-Acc}  &&   & \\
  \midrule

  \textsc{AR}  & 66.2 \small $\pm 0.3$ & 4.4 \small $\pm 0.3$ & 43.4 \small $\pm 1.4$ & 38.0 \small $\pm 0.5$  && 0.716  \small $\pm 0.004$ & 34.5  \small $\pm 0.4$ && - & - \\ % exp-13
  
    \addlinespace[0.2cm]
  
 \multicolumn{3}{l}{\textit{ \dual roll-in }} \\
   \textsc{\dualtarget} & 66.1 \small $\pm 0.2$  & 2.2 \small $\pm 0.2$  & 45.5 \small $\pm 1.2$  & 37.9  \small $\pm 0.4$ && 0.656 \small $\pm 0.003$ & 29.7 \small $\pm 0.2$ && 50K & 1.175 \\  % exp-9
   \textsc{\dualsource} & 66.5 \small $\pm 0.1$ & 3.6 \small $\pm 0.2$ & 49.3 \small $\pm 0.5$ & 39.8 \small $\pm 0.2$ && 0.733 \small $\pm 0.003$ & 37.7 \small  $\pm 0.4$ && 10K & 2.669 \\  %exp-8
  
  \addlinespace[0.2cm]
  
 \ourrollin & 66.1 \small $\pm 0.2$ & \bf{5.2}  \small  $\pm 0.1$& 51.7  \small $\pm 0.2$ & 41.0 \small $\pm 0.1$ &&0.745  \small$\pm 0.005$ & 39.7 \small $\pm 0.2$ && 6K & 2.161 \\        % exp-119
  \frameshort  & 66.8  \small $\pm 0.2$  & 4.9   \small$\pm 0.2$ & \bf{53.3}  \small $\pm 0.4$  & \bf{41.7}  \small $\pm  0.3$ && \bf{0.747}  \small $\pm 0.004$ & \bf{39.8} \small $\pm 0.3$ && 12K & 1.802 \\
  %exp-118
  \bottomrule
 \end{tabular}
 }
\caption{Results on the Newsela-Grade test dataset for Controllable Simplification: our proposed framework, \frameshort, achieves the best performance on \sari and \ari-based metrics across the board. }\label{tab:main_refine_simp} 

\end{table*}

% for table positioning

% summarization main
\begin{table*}[t]
\centering
 \setlength\tabcolsep{4pt}
\scalebox{0.90}{
\begin{tabular}{lrrrrrrrr}
 \toprule
  \multirow{2}{*}{\textbf{Model}} & \multicolumn{4}{c}{\bf{SARI}}  &  &  \multicolumn{3}{c}{\textbf{Rouge-L}} \\
 \cmidrule{2-5}  \cmidrule{7-9}
 &   \textbf{keep-F1} & \textbf{add-F1} & \textbf{del-P} & \textbf{combined} && P & R & F1   \\
  \midrule
  \textsc{AR}  &  20.0 & 1.7 & 58.5 & 26.8 && 35.6 & 30.1 & 32.1\\
    \addlinespace[0.2cm]
 \multicolumn{3}{l}{\textit{ \dual roll-in }} \\
  \textsc{\dualtarget} & 49.5 & 3.7 & 58.8 & 37.3 && 54.2 & 70.1 & 60.8   \\ 
  \textsc{\dualsource}  & 45.5 & 3.6 & 61.4 & 36.8&& 52.8 & 63.4 & 57.2 \\
  \addlinespace[0.2cm]
  \ourrollin & 54.7 & 4.1 & 62.9 & 40.6  && 55.9 &\bf{ 74.6} & 63.6   \\   
  \frameshort & 54.4 & 4.4 & 65.5 & 41.4&& 56.1 & 74.0 & \bf{63.8} \\ 
  \bottomrule
 \end{tabular} 
 }

\caption{Results on the Summarization dataset: \frameshort improves \rouge-\fmeasure and \sari over \editor.}\label{tab:main_refine_summ}
 \vspace{-0.5cm}
\end{table*}

\section{Findings} \label{sec:main_results}

\paragraph{Controllable Simplification} As can be seen in Table~\ref{tab:main_refine_simp}, our overall training framework, {\ttfamily \frameshort} improves over the prior training strategy for \editor  --- {\ttfamily \dualtarget} --- significantly for all metrics (SARI: +3.8, PCC:  +0.091, ARI-Acc: +10.1\%), and over the AR baseline. Ablations show that this is a combined effect of multiple factors. Dual-path roll-in, {\ttfamily \dualsource} improves over {\ttfamily \dualtarget} as expected (\sari: +1.9, PCC:  $+0.077$, ARI-Acc: +8.0\%), as the roll-in sequences encountered during training are similar to those encountered during inference. Using expert roll-in ({\ttfamily \ourrollin}) performs better than using learned roll-in (\dual roll-in) across the board, with gains of up to $3$ SARI points over {\ttfamily \dualtarget}.  Training with \cl ({\ttfamily \frameshort}) improves over the best roll-in strategy\footnote{As the order of the training samples as governed by our curriculum strategy will be same for {\ttfamily \dualsource, \ourrollin}, we only report results over the best roll-in strategy.}, improving the precision of deletions (+1.6) and leading to a significant improvement in SARI score (+0.7) over {\ttfamily \ourrollin} with no significant change in grade-specific metrics.

We also report training and inference statistics. For training, we report the number of training updates to convergence, i.e. when the model achieves the best validation perplexity on the development dataset. For inference, we report the average number of actions taken by the model to generate the refined output counts. Each iteration encompasses a reposition operation followed by an insertion applied to the all the tokens in the input sequence in parallel. \cl reduces the average number of actions needed to generate outputs compared to {\ttfamily \ourrollin}, while taking only $\sim2$K more updates during training than {\ttfamily \dualsource}. These results show that our roll-in policy, \ourrollin and the curriculum play a complementary role in improving training for editing.

\paragraph{Abstractive Summarization}

On the Abstractive Summarization task (Table~\ref{tab:main_refine_summ}), {\ttfamily \frameshort} achieves the best performance across the board compared to alternative training strategies for \editor with gain of upto $\sim4$ \sari, and $\sim3$ \rouge points. Our proposed approach improves the precision of the deletion operation \textsc{(del-P, $+7$)}. It also preserves the tokens from the source sequence that are present in the reference suggested by the improvement in \textsc{keep-f1}($+3.9$) over the \editor ({\ttfamily \dualtarget}) model.

For completeness, we also compare our approach with systems trained in prior work:  
(1) \textsc{ILP} \cite{clarke2008global}, an integer linear programing approach for deletion-based compression, (2) \textsc{T3} \cite{cohn-lapata-2008-sentence}, a tree transducer-based model for abstractive compression, (3) \textsc{Seq2Seq} \cite{filippova2015sentence}, a neural network model for deletion-based compression, (4) \textsc{NAMAS} \cite{rush2015neural}, a neural model for abstractive compression and summarization and (5) \felix \cite{mallinson-etal-2020-felix}, a non-autoregressive approach to text editing. We use the outputs provided by \citet{toutanova-etal-2016-dataset} for [1-4] and \citet{mallinson-etal-2020-felix} for [5]. We endeavored to make the comparison as fair as possible\footnote{We detokenized and manually checked the outputs from \citet{mallinson-etal-2020-felix} and corrected for de-tokenization errors such as ``1. 23'' to ``1.23'' and ``wanda 's'' to ``wanda's''.}, but it is not possible to have a fully controlled comparison. In particular, \felix is trained on uncased data and generates uncased outputs, while we train and evaluate our models with truecasing.

 When evaluated using our pipeline, our training strategy applied to generic \nar models achieve scores that are on par with, or better than, those of dedicated summarization models (Table~\ref{tab:main_summ_baseline}). However, this evaluation penalizes \felix as it is trained to address the simpler problem of summarization on uncased text. On lower-cased outputs, our best model falls behind \felix by $1.7$ \rouge points. However, \felix has about twice as many parameters as our model and benefits from BERT pre-training \cite{DevlinCLT19}. As a result, this comparison confirms the promise of our approach overall.

% summarization main
\begin{table}[h]
\centering
\vspace{-0.1cm}
 \setlength\tabcolsep{4pt}
\scalebox{0.90}{
\begin{tabular}{lrrr}
 \toprule
  \multirow{2}{*}{\textbf{Model}} &   \multicolumn{3}{c}{\textbf{Rouge-L}} \\
 \cmidrule{2-4}
  & P & R & F1   \\
  \midrule
  
  \textsc{ILP} \cite{clarke2008global} & \bf{60.6} & 63.2 & 60.6 \\
   \textsc{T3} \cite{cohn-lapata-2008-sentence}& 48.3 & 20.0 & 26.8  \\
   \textsc{NAMAS} \cite{rush2015neural} & 48.8 & 55.2 & 51.5  \\
     \textsc{Seq2Seq} \cite{filippova2015sentence} & 57.6 & 51.5 & 53.1 \\
  \felix \cite{mallinson-etal-2020-felix} & 53.7 & 58.1 & 55.5  \\
  \frameshort & 56.1 & \bf{74.0} & \bf{63.8} \\ 
  \midrule
      \felix\textsc{(lc)} & 65.3 & 71.5 & \bf{67.8}\\ 
 \frameshort \textsc{(lc)}  & 57.7 & 77.2 & 66.1\\ 
  \bottomrule
 \end{tabular} 
 }
\caption{Comparison to prior work on Summarization dataset: Our approach outperforms all the baselines in \rouge-L (\fmeasure). \textsc{LC}:lower-cased.}\label{tab:main_summ_baseline} 
\vspace{-0.7cm}
\end{table}

%

% Section 6: Ablation 

\section{Analysis}

We conduct further experiments to better understand the factors that help our training strategies improve editing quality.

\subsection{Impact of \ourrollin \rollin}

\begin{figure}[h]
\begin{subfigure}{0.40\textwidth}
\centering
  \includegraphics[width=0.48\linewidth]{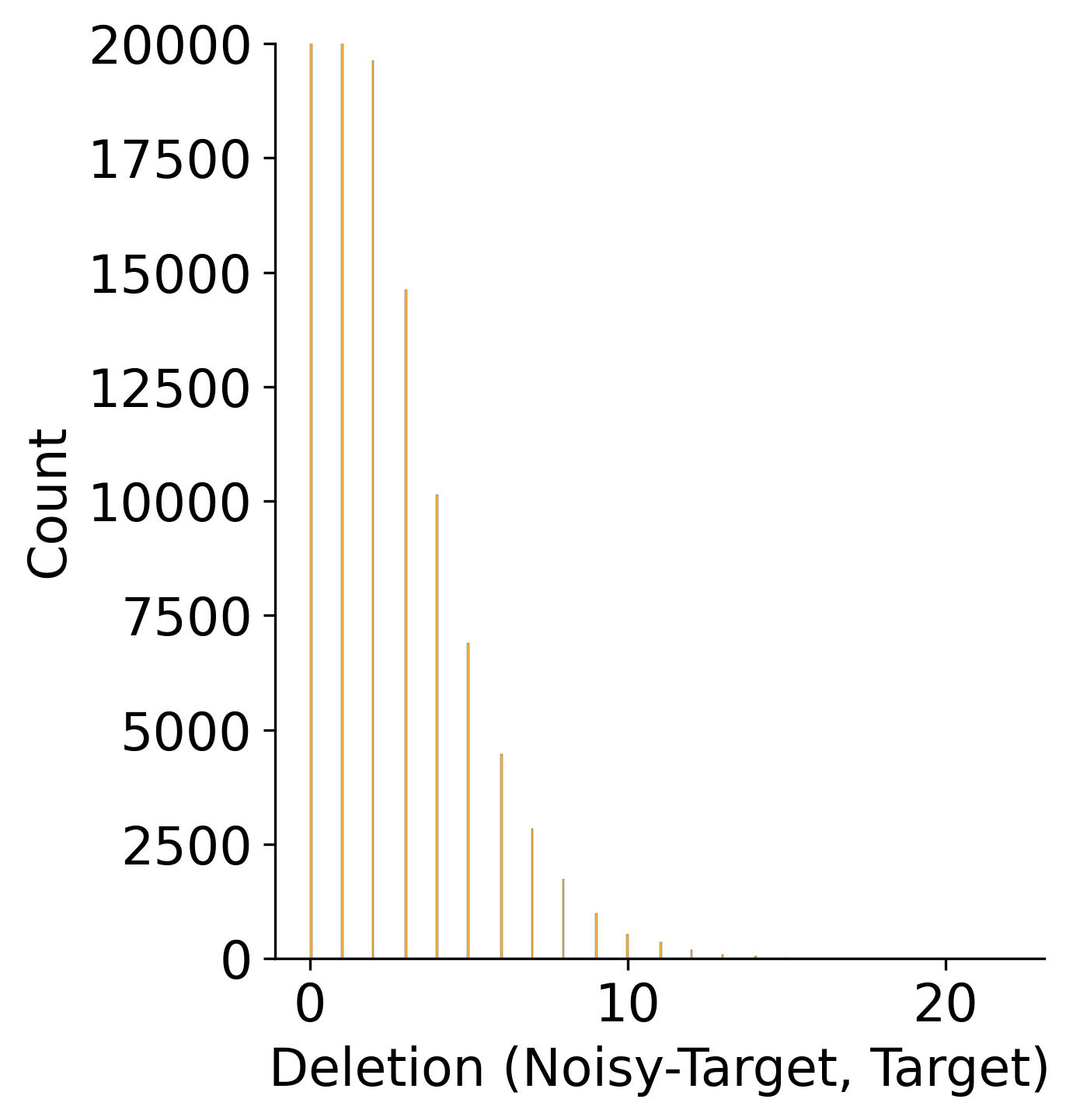}
  \includegraphics[width=0.48\linewidth]{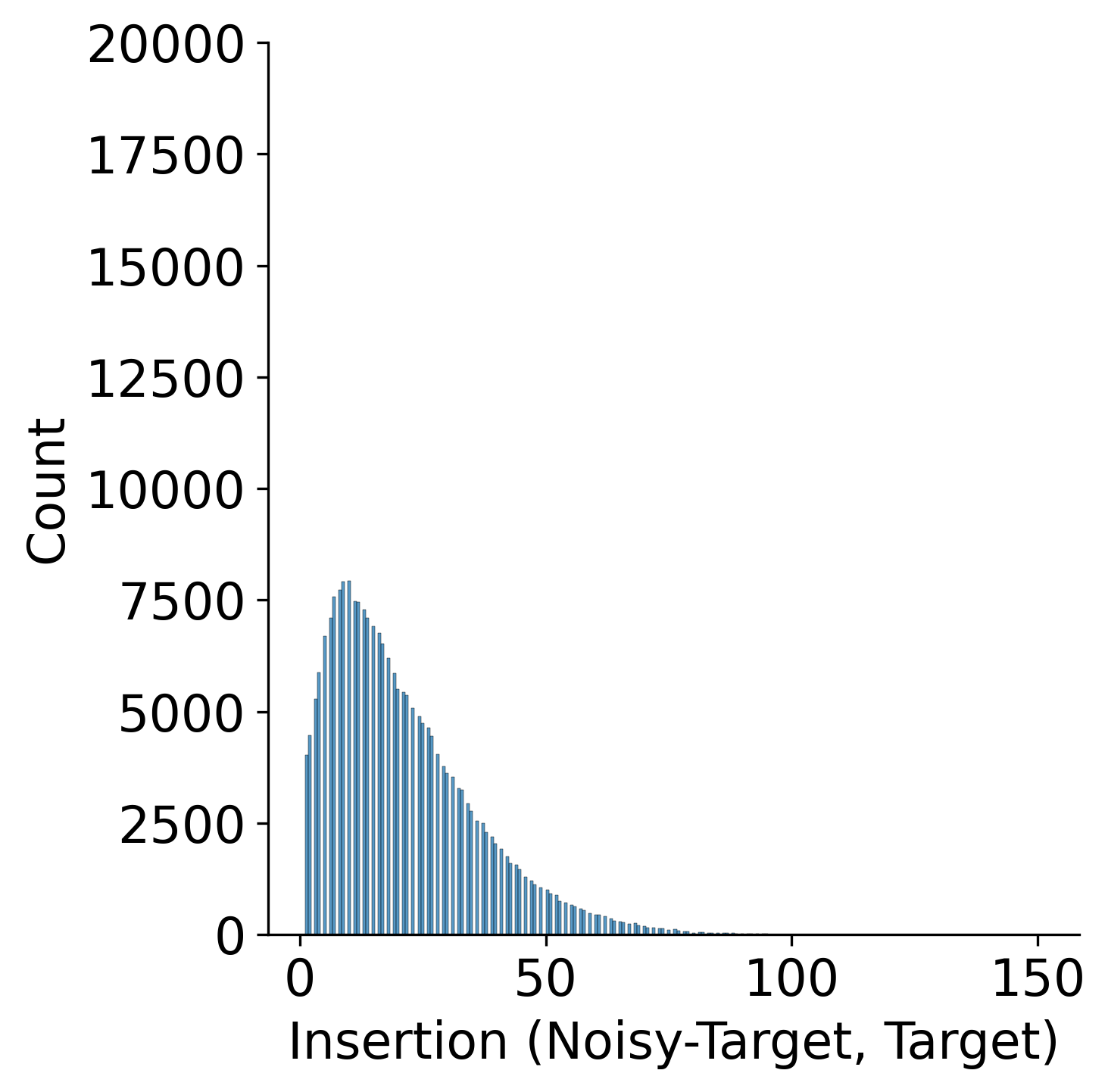}
 \caption{\editor\'s \rollin \small{(Training)}}
\end{subfigure}
% \hspace{0.05cm}
% \rulesep
% \hspace{0.05cm}

\begin{subfigure}{0.40\textwidth}
\centering
  \includegraphics[width=0.48\linewidth]{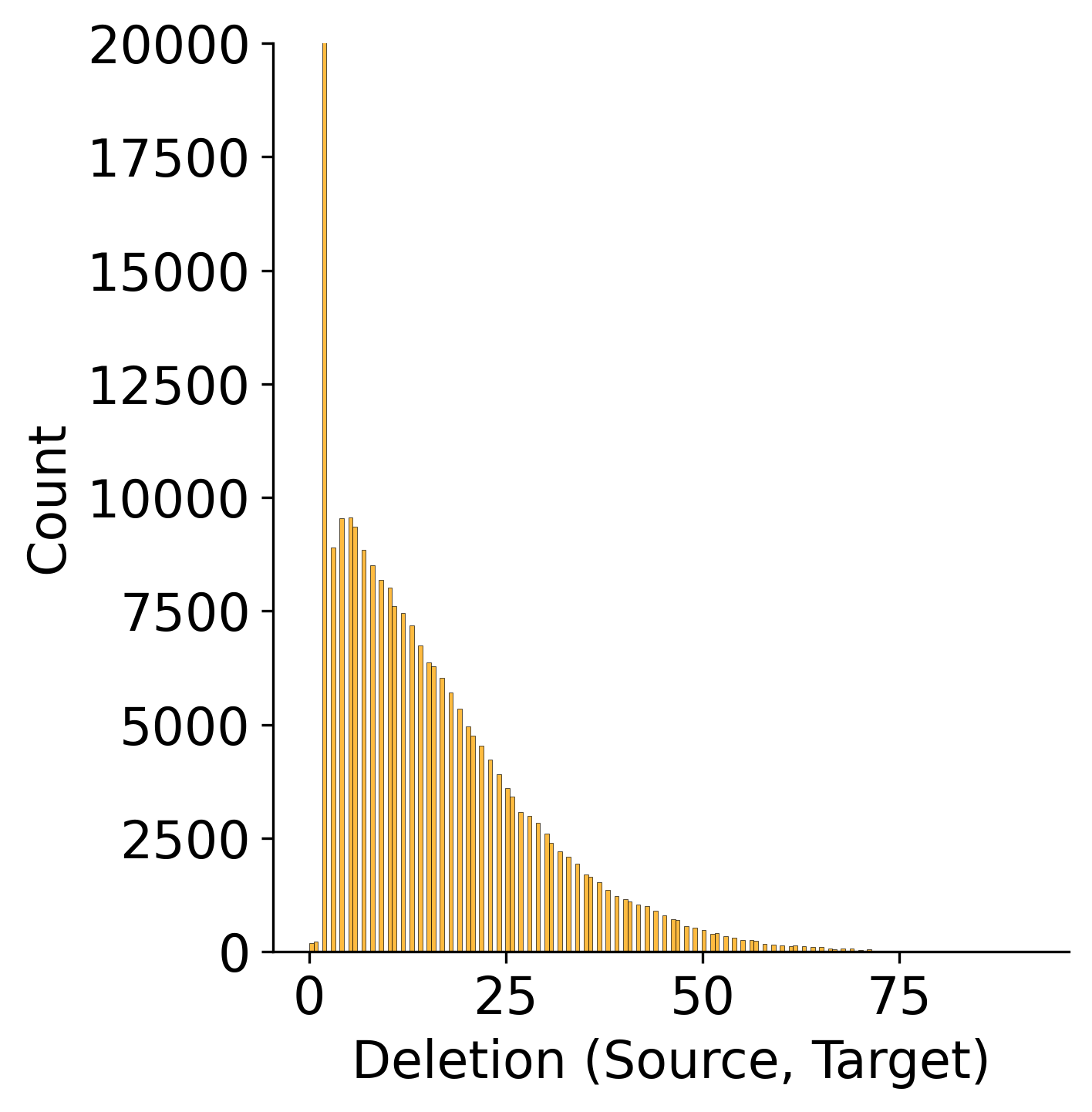}
  \includegraphics[width=0.48\linewidth]{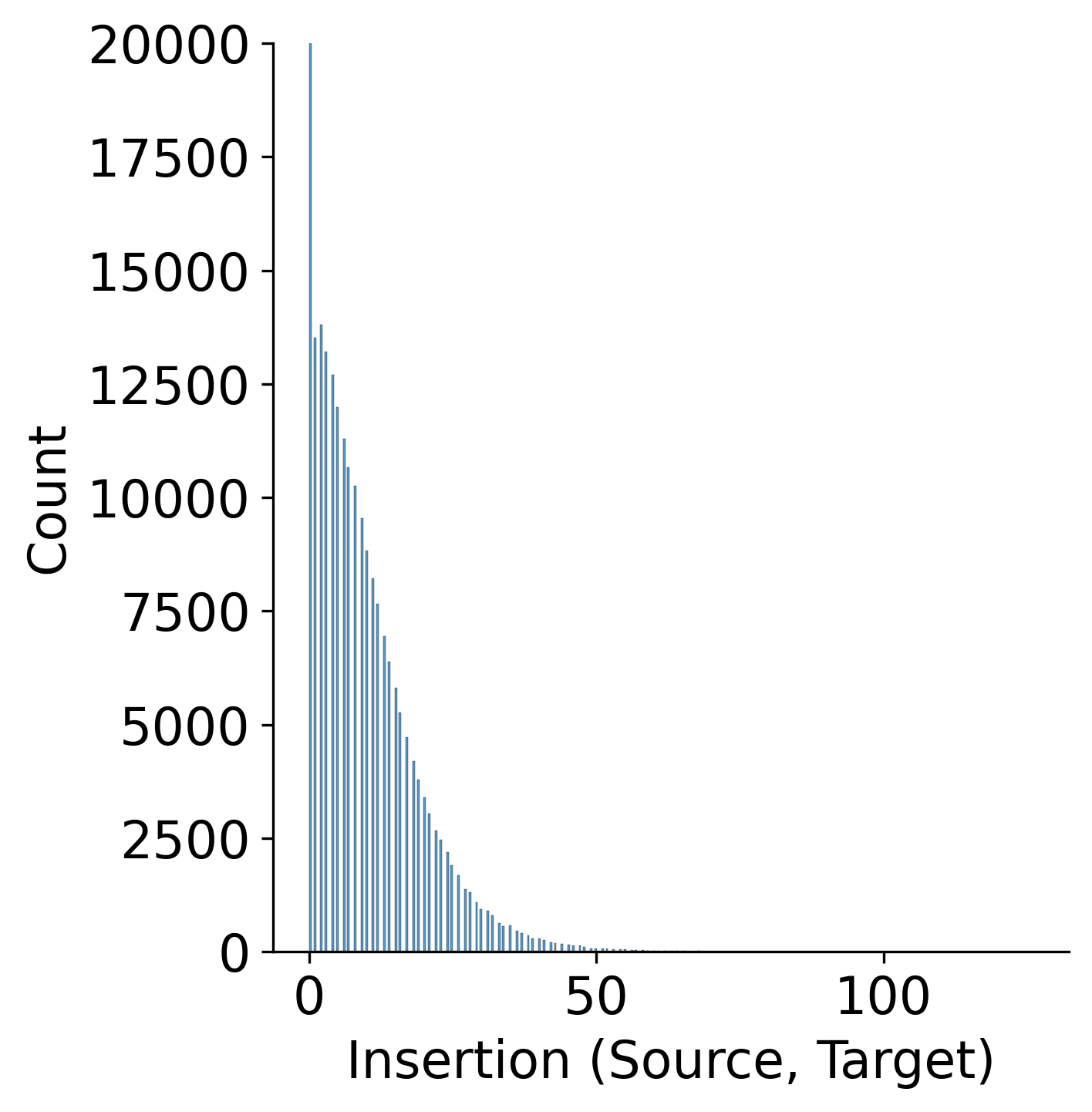}
  \caption{Inference Distribution}
\end{subfigure}
% \hspace{0.05cm}
% \rulesep
% \hspace{0.05cm}

\begin{subfigure}{0.40\textwidth}
\centering
  \includegraphics[width=0.48\linewidth]{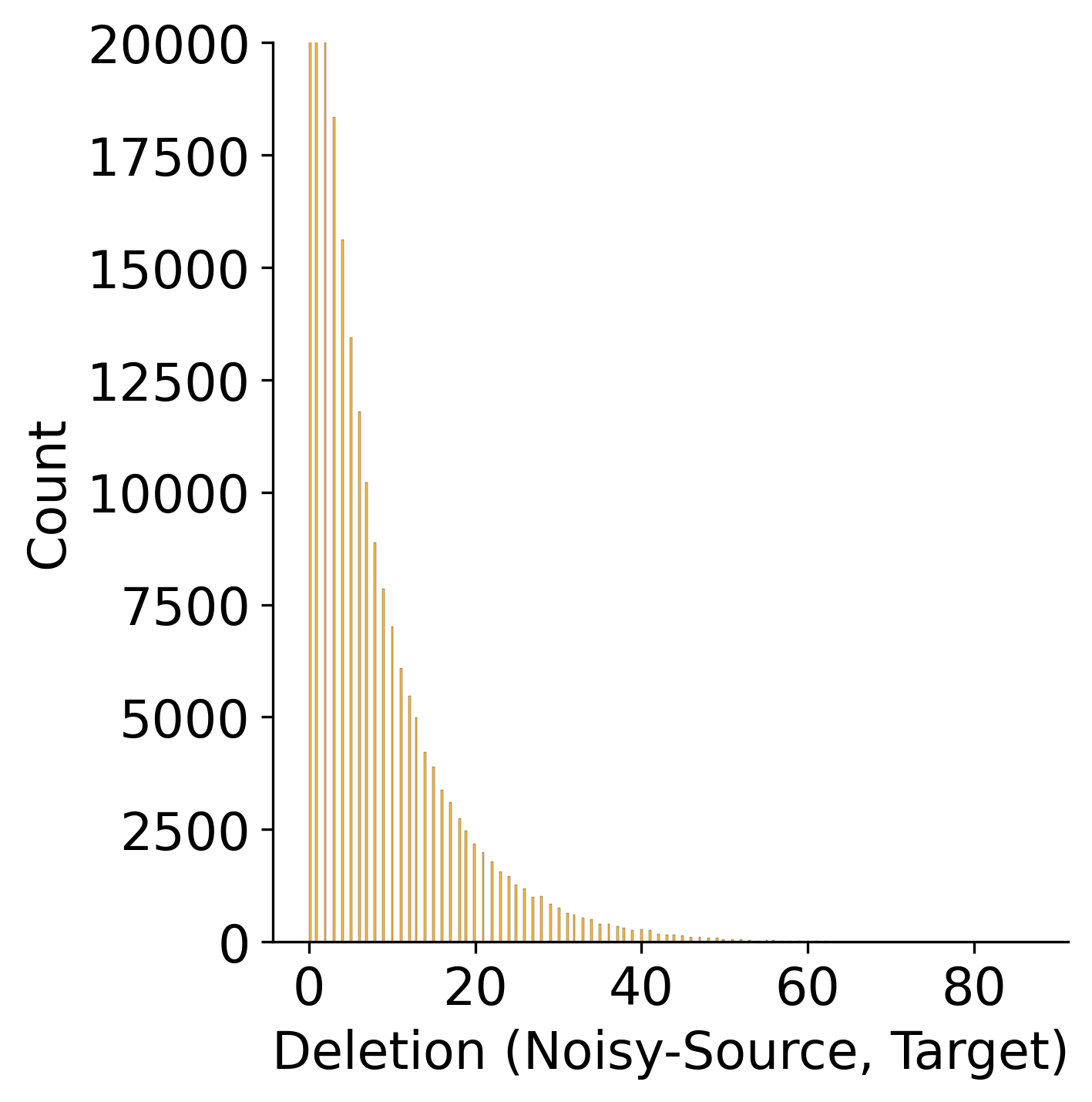}
  \includegraphics[width=0.48\linewidth]{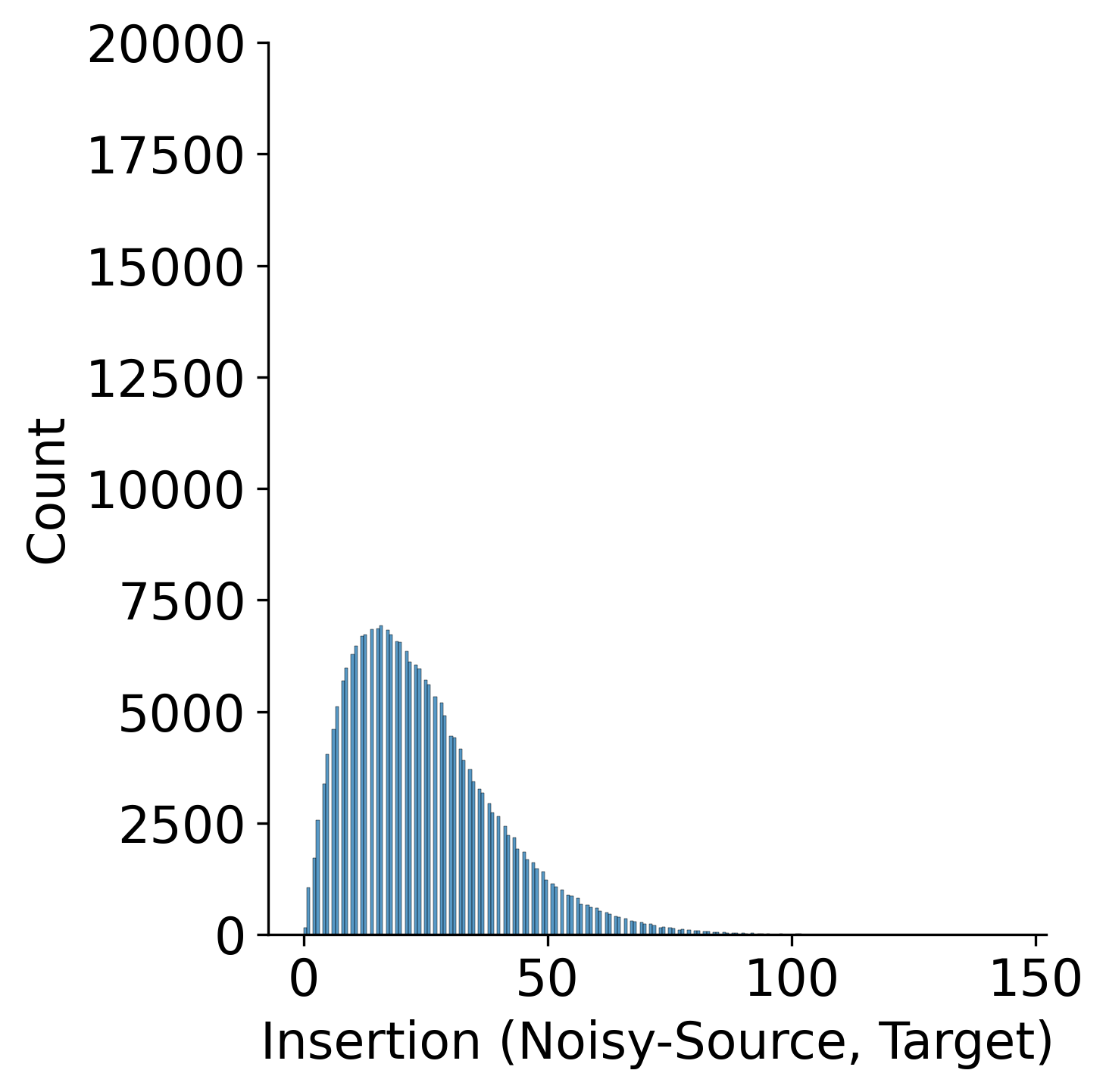}
  \caption{\ourrollin \rollin \small{(Training)}}
\end{subfigure}%

 \caption{ Distribution of Oracle Edit Operations (\textit{\color{blue}Insertions}/\textit{\color{yellow}Deletions}) observed on Controllable \ts. Our proposed \rollin policy\'s distribution of edit operations is closer to the inference distribution, while enabling exploration during training.
 } \label{fig:edit_dist_analysis}
 \vspace{-0.5cm}
\end{figure}

First, we seek to measure whether our approach has the intended effect of bridging the gap between training and test for editing tasks.
Figure~\ref{fig:edit_dist_analysis} shows the distribution of oracle insertion and deletions observed when (a) training with  \editor\'s default \rollin policy; (b) refining an original \source sequence and (c) exposed to the model with our \ourrollin \rollin policy for Controllable \ts. The plots show that with the default learning policy of the Editor model, the model doesn't learn to perform complex deletion operation at inference time. By contrast, our proposed \rollin exposes the model to the distribution that has higher overlap with the inference distribution as as well as additional intermediate sequences that encourages exploration during training.

%%%%%%%%%%%%%%%%%%%%%%%%%%%%%%%%%%%%%%%%%

\begin{figure*}[htb!]
\begin{subfigure}{0.5\textwidth}
\centering
  \includegraphics[width=\linewidth]{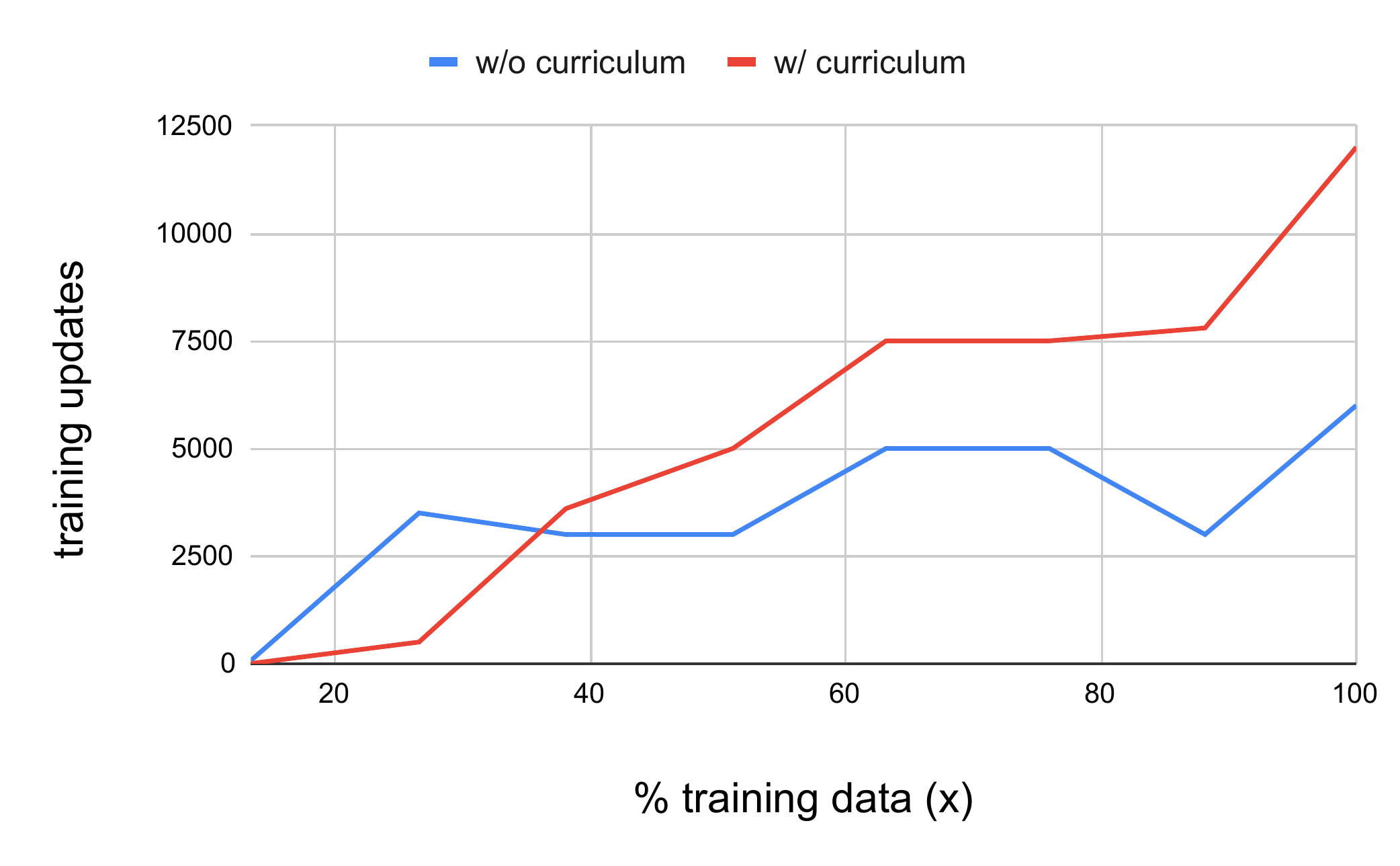}
\end{subfigure}
\begin{subfigure}{0.5\textwidth}
  \centering
  \includegraphics[width=\linewidth]{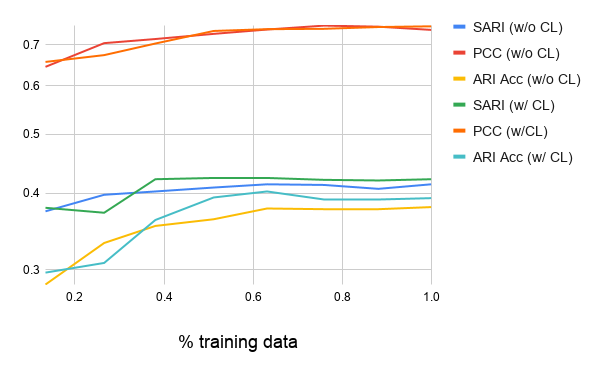}
\end{subfigure}%
\vspace{-0.5cm}
 \caption{ The sample order during training matters as training without curriculum on the same amount of data ($>=40\%$) converges early (plot on the left) and to lower performance across all metrics (plot on the right) relative to training with curriculum using the same data.  }
 \label{fig:percent_data_conv_with_cl }
  \vspace{-0.5cm}
\end{figure*}

\begin{figure}[htb!]
\centering
  \includegraphics[width=0.8\linewidth]{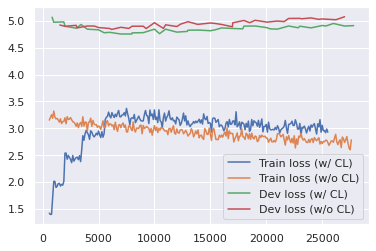}
    \caption{Training with curriculum reduces the loss on the development dataset leading to better generalization on Controllable \ts.}  \label{fig:training_cl}
     \vspace{-0.5cm}
\end{figure}

\subsection{Impact of Curriculum Controlled \rollout} 

\paragraph{Training Dynamics} To verify that curriculum learning helps our model better exploit its training data, we train \editor on $x\% \in [0,100]$ of the data, and compare using random samples with samples ranked by increasing edit distance. Figure ~\ref{fig:percent_data_conv_with_cl } shows the number of updates to convergence on the development dataset for controllable simplification with/without \cl. Training converges early (70 iterations only) on ~13\% of the easiest samples with oracle edit distance between the \source and the \target sequence $<=2$. This supports the hypothesis that despite adding noise, our approach yields easier examples to train on. The order in which samples are presented matters, as adding batches with larger edit distance  ($>63$\% data) without maintaining the order of the samples converges early. By contrast, the curriculum pacing function adds samples in order of increasing difficulty, allowing the model sufficient training time to learn from new samples while improving overall performance across metrics. 

We also report the learning curves when training \editor on the Newsela dataset in Figure~\ref{fig:training_cl}. Training with curriculum reduces the overall loss consistently on the development dataset, leading to better generalization.

% simplification main
\begin{table}[h]
\centering
 \setlength\tabcolsep{4pt}
\scalebox{0.85}{
\begin{tabular}{lrrr|r}
 \toprule
 \textbf{Criteria} & \textbf{SARI}& \textbf{PCC} &  \textbf{\% ARI-Acc} & \textbf{Corr.} \\
\midrule
Random & $40.7$ & $0.749$ & $38.6$ & - \\
Length Ratio & $41.0$ & $0.762$ & $39.0$ & $0.26$ \\
Grade Difference & $40.7$ & $0.730$ & $38.3$ & $0.19$\\
\midrule
\frameshort & $42.0$ & $0.758$ & $39.6$ & $1.00$ \\
- \ourrollin \rollin & $40.1$ & $0.734$ & $37.8$ & - \\
- \cl & $41.2$ & $0.742$ & $39.3$ & - \\
  \bottomrule
 \end{tabular}
 }
\caption{On Newsela-grade dev dataset: Using Edit distance as the difficulty criteria improves over both task-specific (Grade Difference) and task-agnostic (Length ratio) criteria. Our proposed \ourrollin \rollin and curriculum-controlled \rollout  provides complementary advantages to the model training.}\label{tab:ablation_curr} 
 \vspace{-0.5cm}
\end{table}

\paragraph{Ranking Criteria} We compare the edit-distance (\frameshort) with other curriculum criteria in Table~\ref{tab:ablation_curr} where the order of examples is a) random, b) controlled by the length ratio between source and target sequence (Length Ratio), c) governed by the difference between the source and target grade levels (Grade Difference).  Our proposed criterion outperforms both task-specific ({\ttfamily Grade Difference}) and task-agnostic criteria ({\ttfamily Length Ratio}) on the Newsela Grade development set across all the metrics. {\ttfamily Length Ratio} achieves better correlation with Edit distance than {\ttfamily Grade Difference} which is also reflected by its performance (\sari: +0.3, \pcc: 0.032, \ari: 0.7)  on the Controllable Simplification task. This might reflect the fact that higher grade differences do not necessarily require more edits to be performed, for instance when the sentence to be simplified is already relatively simple. These mismatches do not occur when the edit distance itself is used as the sample difficulty criterion.

% \sa{Added to include without cl results}
\paragraph{Complementarity of \rollin and \rollout design} We report the performance of the {\ttfamily \dualsource} model, when trained with curriculum only without the \ourrollin policy, i.e.  {\ttfamily \frameshort - \ourrollin} in the same Table~\ref{tab:ablation_curr}. Both {\ttfamily \ourrollin} \rollin and curriculum controlled roll-out provides complementary advantages to the model training as removing either results in the drop in performance across all the metrics for controllable \ts. However, we observe larger drop in the scores when we do not apply the \ourrollin policy which shows that our proposed \rollin policy is necessary to reap the benefits of curriculum learning.

%

% Section 7: Related Work 
\section{Related Work}

% NAR edit based models
\paragraph{NAR models} They have been used to enable parallel generation of output tokens for Machine translation. \cite{stern2019insertion, chan2020imputer, xu2021editor}. \citet{ mallinson-etal-2020-felix} design a custom multi-step non-autoregressive edit-based model for sequence editing where each source token is first tagged to represent the type of edit operation to be performed and then a secondary model is used to in-fill new tokens. The tagging and editing models are trained independently. By contrast, we propose approaches to adapt \nar models designed for MT for these tasks and train an end-to-end model to generate an edited sequence. 

% Curriculum learning 
\paragraph{Curriculum Learning for Sequence Refinement} While curriculum learning has been applied to many tasks such as MT \cite{haffari2009machine, platanios2019competence, kumar-etal-2019-reinforcement}, sentiment analysis \cite{sido2019curriculum}, natural language understanding \cite{xu2020curriculum}, reading comprehension \cite{tay2019simple}, their application to sequence refinement tasks has not been explored yet. Various strategies have been proposed to control the sample difficulty like n-gram frequency \cite{haffari2009machine, platanios2019competence}, token rarity, and sentence length \cite{liu2020norm}. \citet{chang-etal-2021-order} use Levenshtein edit distance as a sample difficulty criteria to order the samples for the task of data-to-text generation where the training model uses an \ar  seq2seq model. Instead, we focus on edit distance as a sample difficulty criteria that is directly tied to the training oracle and model design. 

% Imitation learning
\paragraph{Roll-in policies} There has been a plethora of work in the Imitation learning landscape on algorithms that strike a balance between learned and expert roll-in policies \cite{ross2011reduction,venkatraman2015improving,chang2015learning}. However, large differences in expert and learner's policy action can hurt performance \cite{brantley-etal-2019-non, he2012imitation, leblond2018searnn}. In our work, we propose to roll-in with noised states instead, so that the model can be exposed to mimic expert demonstrations from states that the model is more likely to encounter during inference.

%

% Section 8: Conclusion
\section{Conclusion}
This paper introduced two complementary strategies to address undertraining and poor generalization when adapting \nar models to editing tasks: 1) a new \rollin policy that generates intermediate sequences that the model is likely to encounter during inference and 2) a curriculum to control the difficulty of the roll-out policy which estimates the cost-to-go from the \rollin sequences to the desired \target sequences, throughout training. Together, these strategies improve output quality consistently on controllable  simplification and abstractive summarization. These results open space for further research to evaluate the potential of this approach for other editing tasks (e.g., post editing, style transfer), and to further tailor imitation learning policies and curriculum design to these tasks.

\section*{Acknowledgments}

We thank Eleftheria Briakou, Khanh Nguyen, Kianté Brantley, the members of the \textsc{CLIP} lab at \textsc{UMD}, and the anonymous \textsc{ARR} reviewers for their helpful and constructive comments.

\bibliography{anthology,references}
\bibliographystyle{acl_natbib}

\appendix
\newpage
\onecolumn
\section{Results on Development set}
% simplification main
\begin{table}[h]
\centering
 \setlength\tabcolsep{2.5pt}
\scalebox{0.80}{
\begin{tabular}{lrrrrcrrcr}
 \toprule
  \multirow{2}{*}{\textbf{Model}} & \multicolumn{4}{c}{\bf{SARI}}  & & \multicolumn{2}{c}{\bf{ARI-based}} & &  \multirow{2}{*}{ \shortstack{\textbf{Inference } \\\textbf{ action/sample }}} \\
 \cmidrule{2-5} \cmidrule{7-8} 
 &   \textbf{keep-F1} & \textbf{add-F1} & \textbf{del-P} & \textbf{combined} & & \textbf{PCC} &  \textbf{ARI-Acc}  &  & \\
  \midrule

  \textsc{AR}  & 0.653 & 0.043 & 0.456 & 0.384 && 0.711 & 0.349 && -\\ % exp-13
  
    \addlinespace[0.2cm]
  
 \multicolumn{3}{l}{\textit{ \dual roll-in }} \\
  \textsc{\dualtarget} &0.648   & 0.021   & 0.454   & 0.374   & & 0.645   & 0.285  & & 1.188  \\  % exp-9
  \textsc{\dualsource} & 0.660   & 0.035   & 0.510   & 0.402 &  & 0.727   & 0.368   && 2.545  \\  %exp-8

\addlinespace[0.2cm]
  
  \ourrollin & 0.657   & 0.049   & 0.530  &  0.412  &  & 0.742   & 0.393 &  & 2.071   \\    
  \frameshort  & 0.662   & 0.043   & 0.556   & 0.420  & & 0.758   & 0.397   && 1.771 \\
  %exp-118
  \bottomrule
 \end{tabular}
 }
\caption{Results on the Newsela-Grade development dataset for Controllable Simplification: our proposed framework, \frameshort, achieves the best performance on \sari and \ari-based metrics across the board.  }\label{tab:dev_refine_simp} 
\end{table}

\section{Impact of Noise}

\begin{figure}[h]
    \centering
  \includegraphics[width=0.50\linewidth]{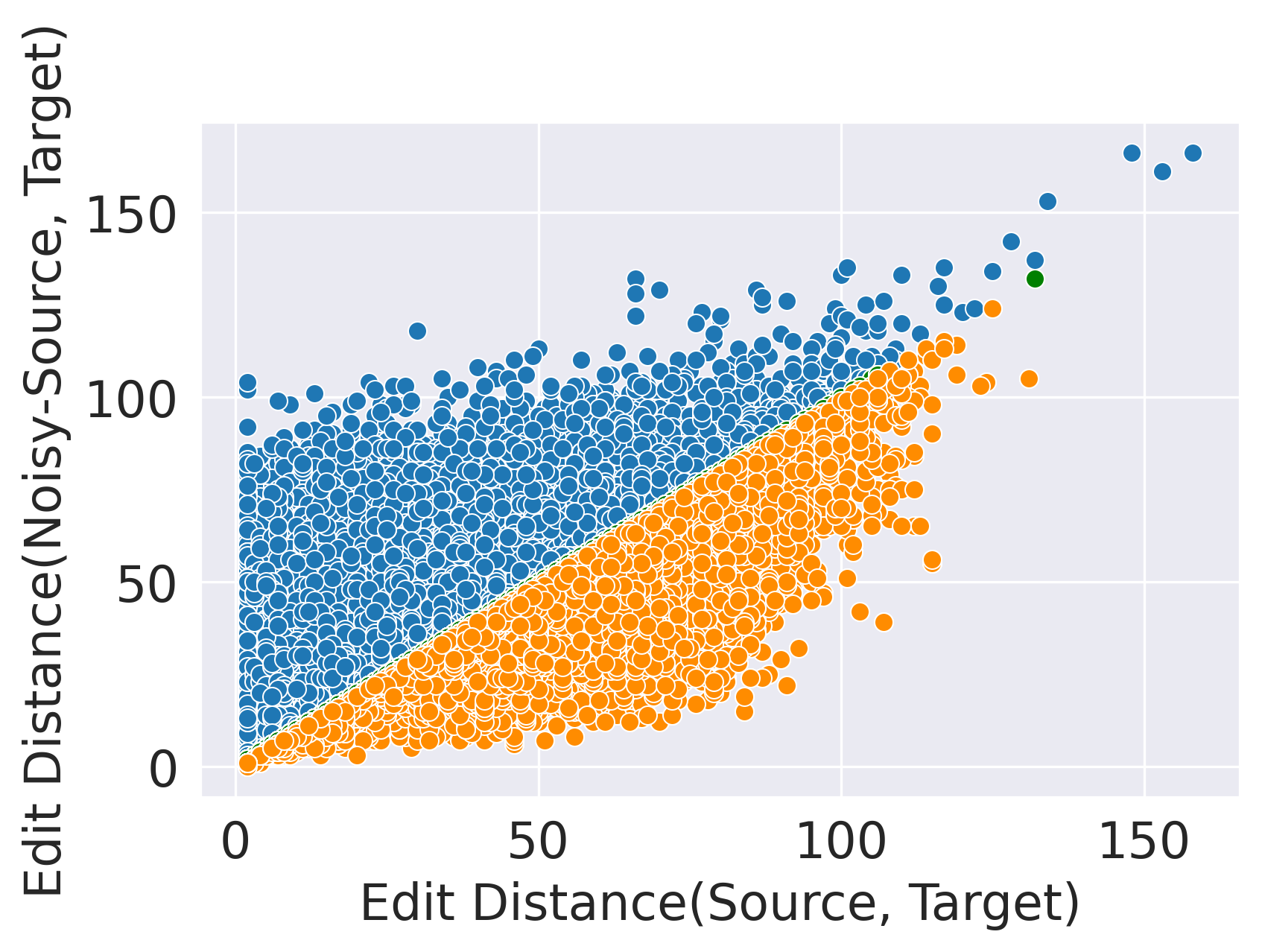}
%   \caption{Summarization}
    \caption{Adding noise to the source increases (\textit{higher}) or decreases (\textit{lower}) the edit distance uniformly across samples for Controllable \ts.}
    \label{fig:effect_noise_2}
\end{figure}

Figure~\ref{fig:effect_noise_2} shows that adding noise to the training samples smoothes the distribution across training instances by creating intermediate sequences that have relatively lower (or higher) overall edit distance with the reference sequence compared to the original \source sequence.

\newpage
\section{Oracle Edit Distribution for Summarization}
\begin{figure}[h]
\centering
\begin{subfigure}{0.60\textwidth}
\centering
  \includegraphics[width=0.48\linewidth]{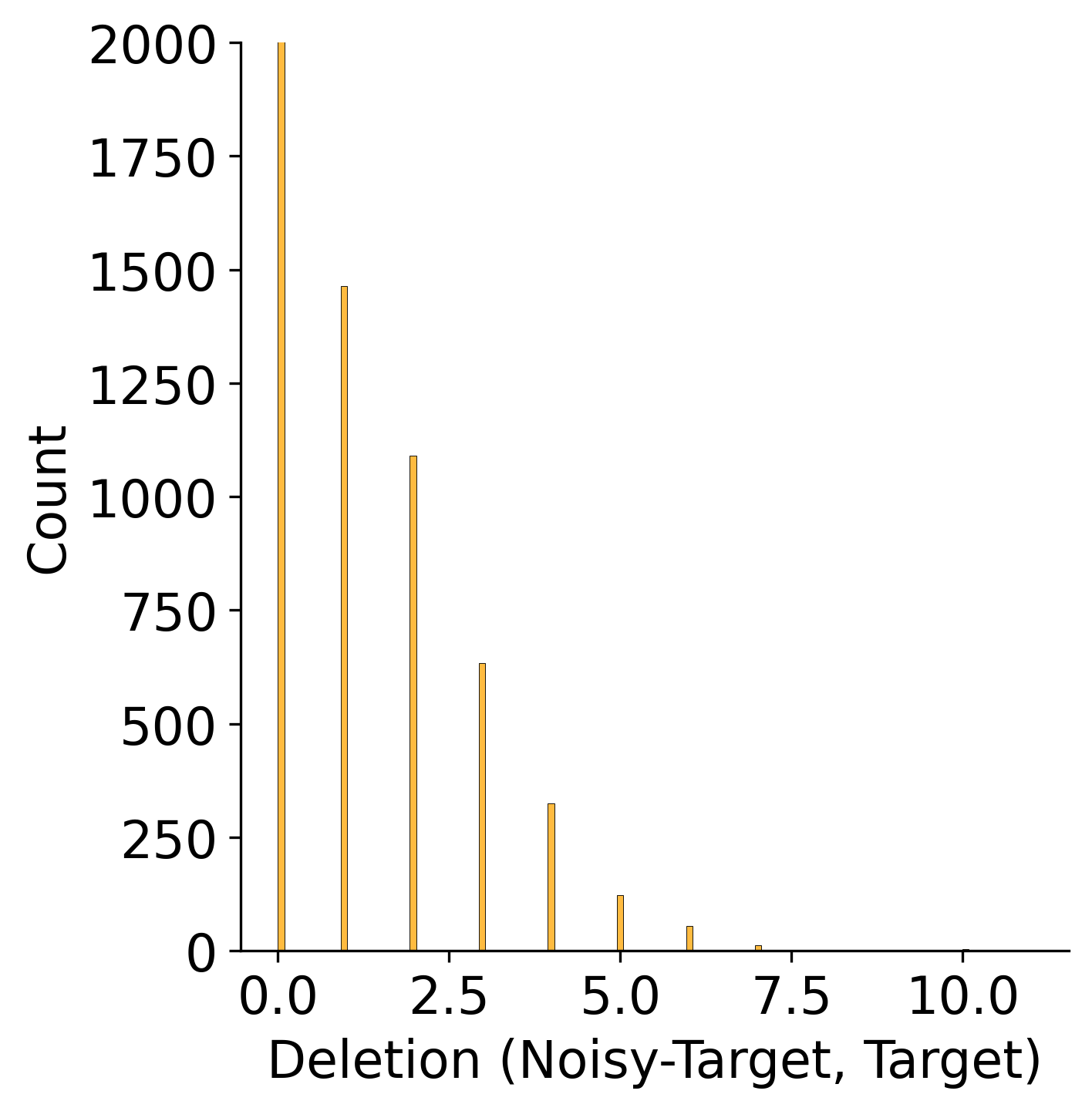}
  \includegraphics[width=0.48\linewidth]{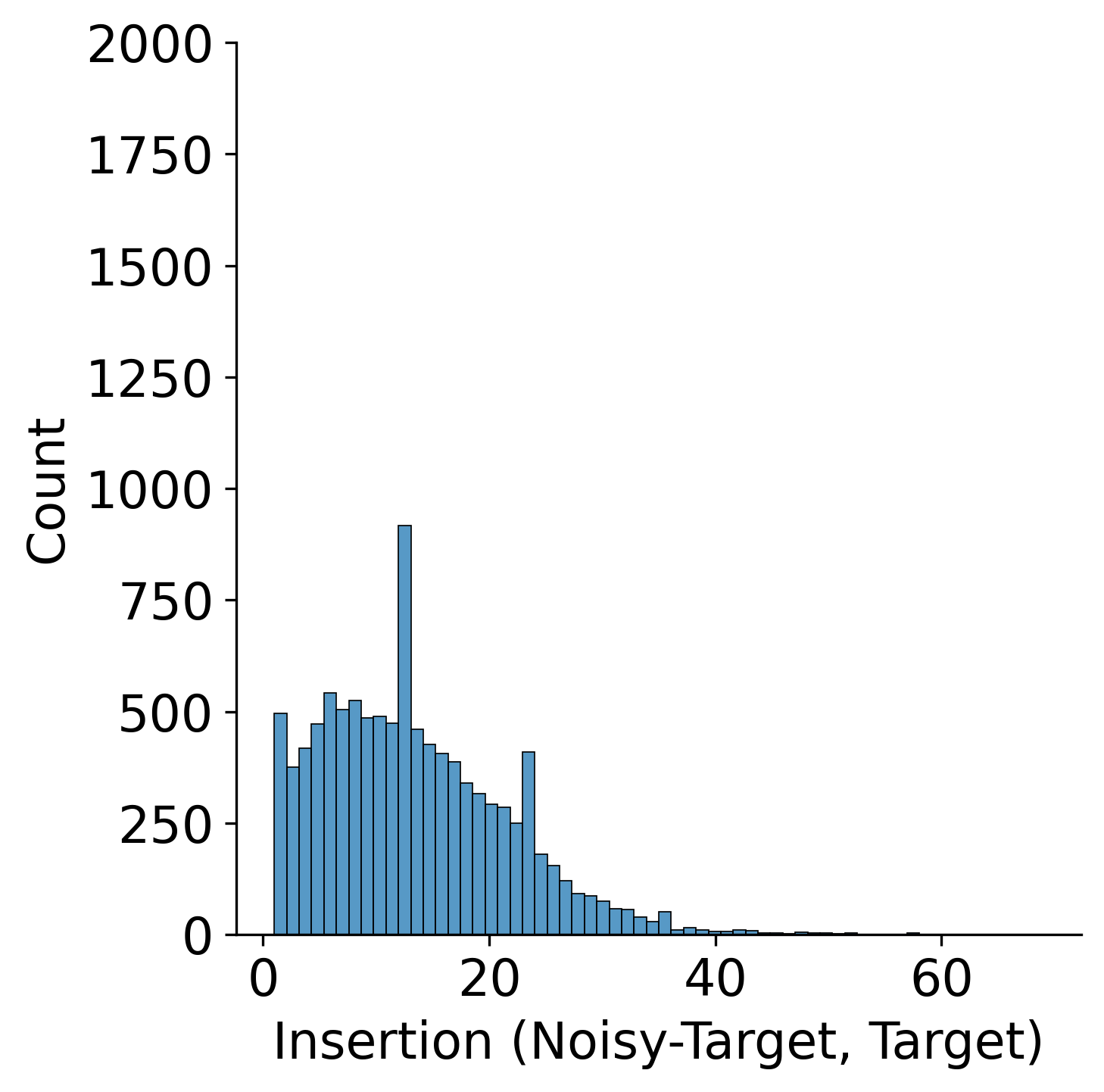}
  \caption{\editor\'s \rollin \small{(Training)}}
\end{subfigure}

\begin{subfigure}{0.60\textwidth}
\centering
  \includegraphics[width=0.48\linewidth]{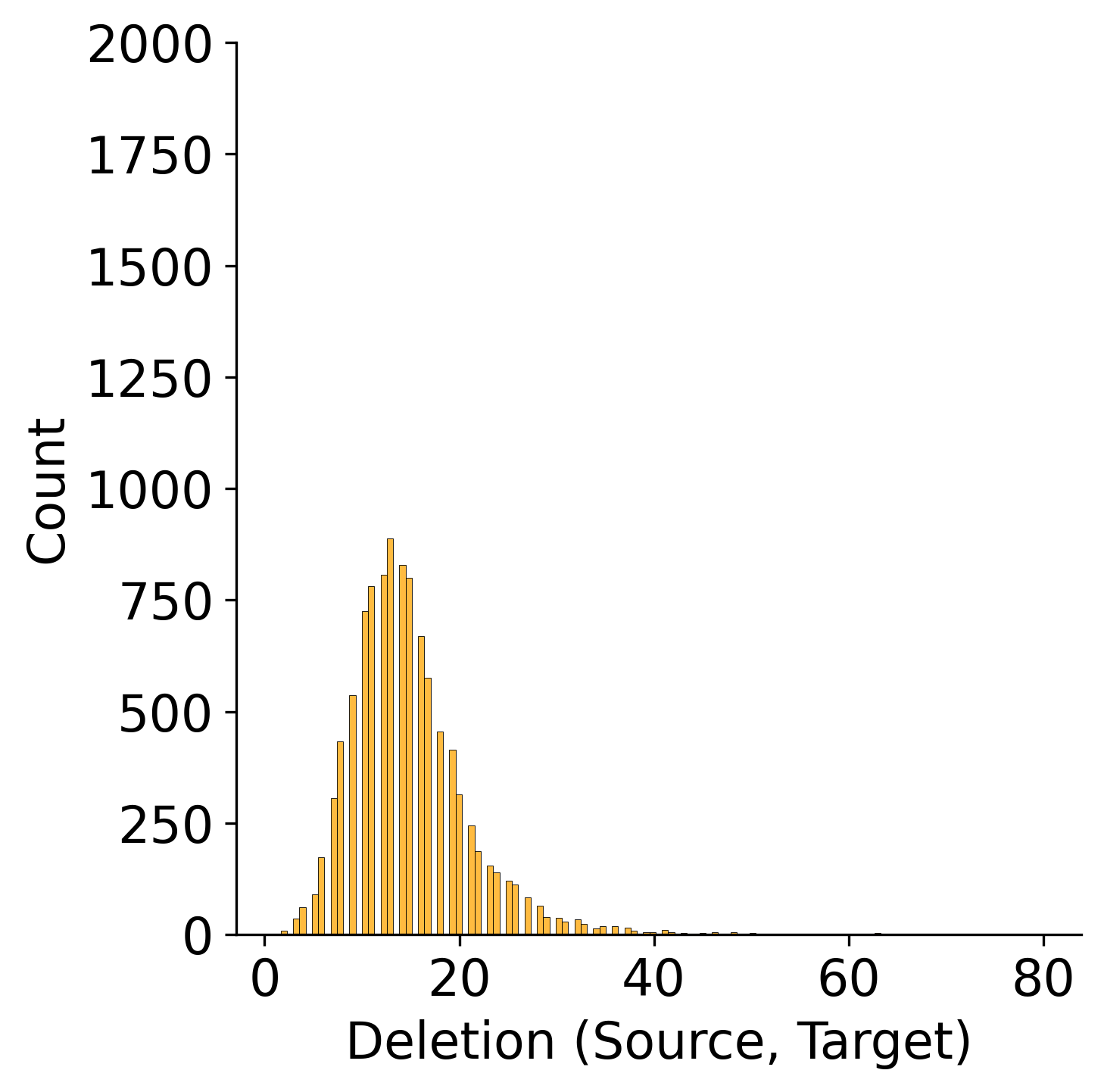}
  \includegraphics[width=0.48\linewidth]{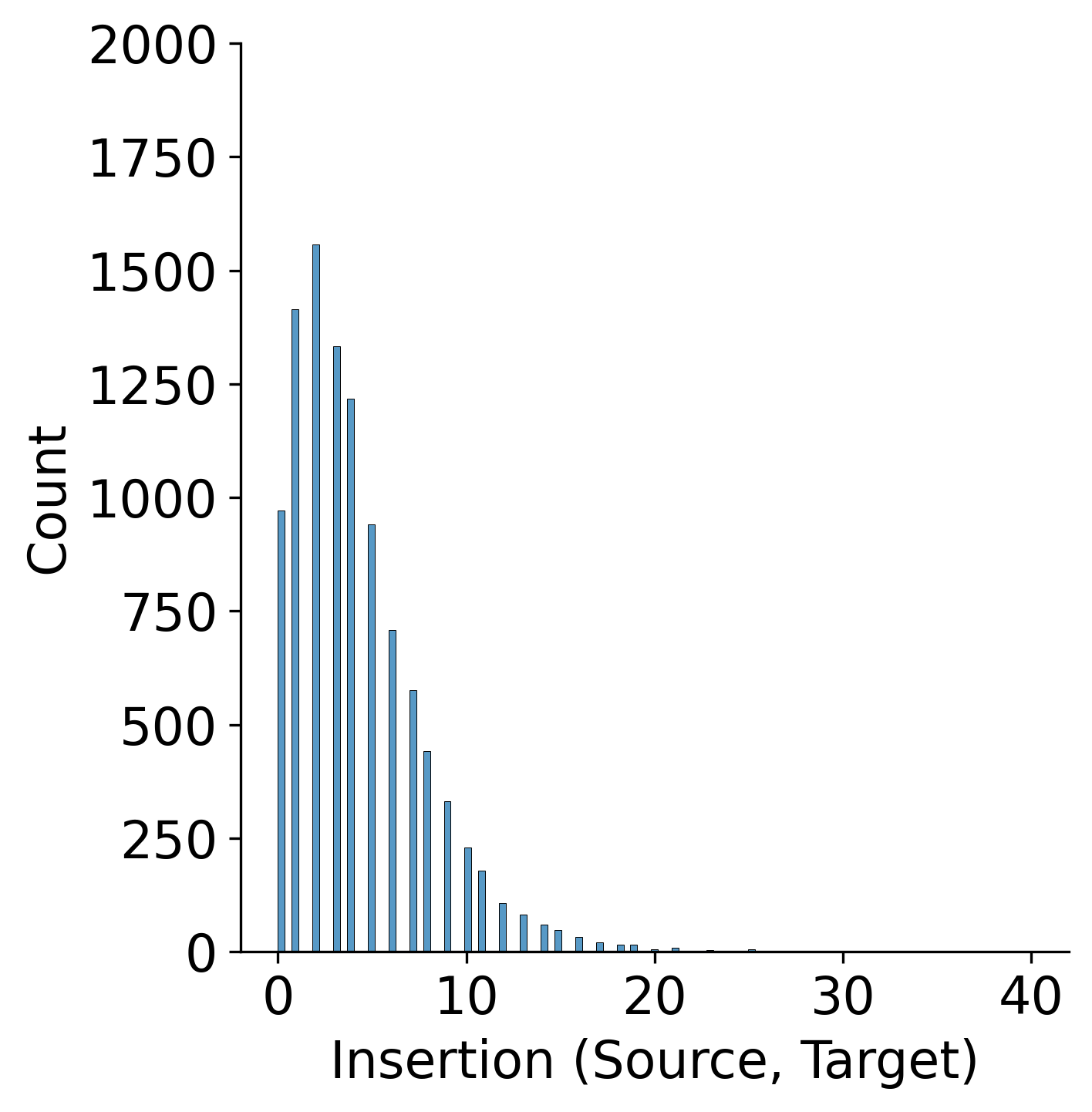}
  \caption{Inference Distribution}
\end{subfigure}

\begin{subfigure}{0.60\textwidth}
\centering
 \includegraphics[width=0.48\linewidth]{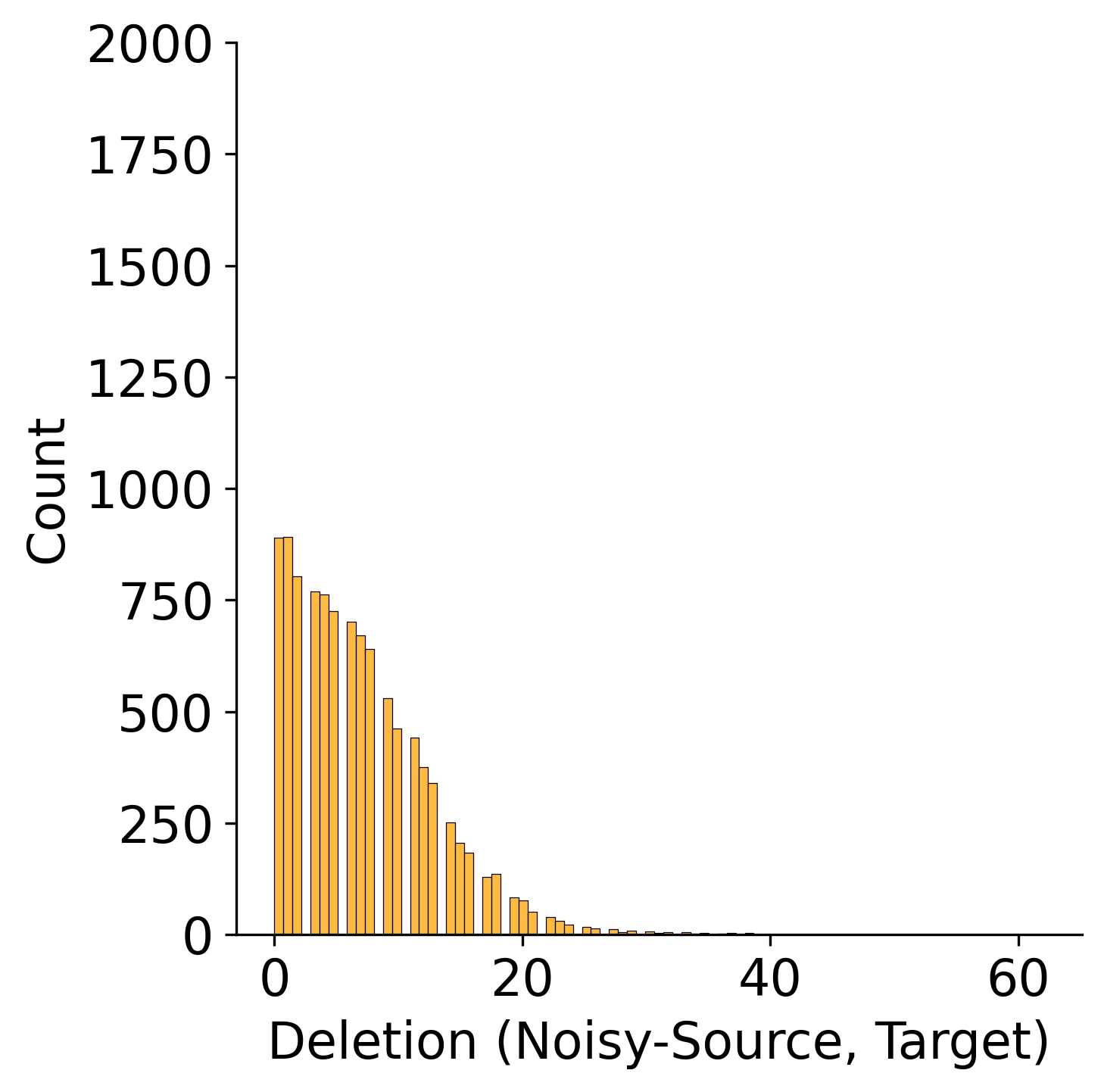}
  \includegraphics[width=0.48\linewidth]{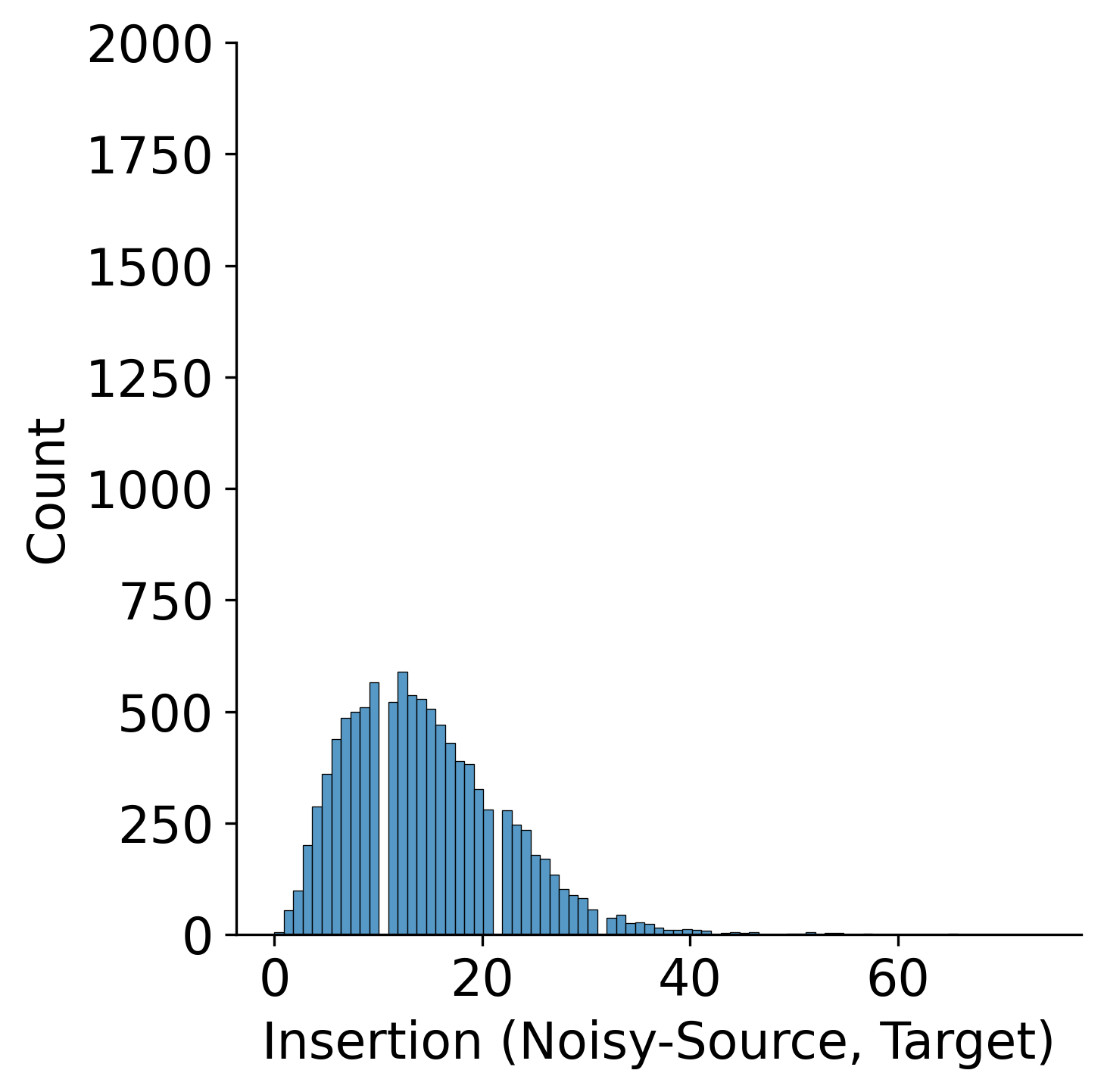}
  \caption{\ourrollin \rollin \small{(Training)}}
\end{subfigure}%
 \caption{ Distribution of Oracle Edit Operations (\textit{\color{blue}Insertions}/\textit{\color{yellow}Deletions}) observed on Abstractive Summarization. Our proposed \rollin policy\'s distribution of edit operations is closer to the inference distribution, while enabling exploration via generated intermediate sequences during training.} 
\end{figure}
% \section{Example Appendix}
% \label{sec:appendix}

% This is an appendix.

\end{document}